\documentclass[12pt,a4paper]{article}

\usepackage[margin=1.1in]{geometry}
\usepackage{amsmath,amssymb,amsthm}

\usepackage{mathtools}
\usepackage{bm}
\usepackage{booktabs}
\usepackage{xcolor}
\usepackage{microtype}
\usepackage{enumitem}
\usepackage{listings}
\lstdefinestyle{pydesign}{
  language=Python,
  basicstyle=\ttfamily\scriptsize,
  breaklines=true,
  columns=fullflexible,
  keepspaces=true,
  showstringspaces=false,
  tabsize=4,
  xleftmargin=1.4em,
  numberstyle=\tiny,
  numbers=left,
  frame=tb,
  framesep=3pt,
  aboveskip=0.5em,
  belowskip=0.2em,
}
\usepackage{tikz}
\usetikzlibrary{arrows.meta,positioning,calc}
\usepackage{pgfplots}
\pgfplotsset{compat=1.18}
\usepgfplotslibrary{groupplots}
\usepackage[numbers,sort&compress]{natbib}
\usepackage{xurl}
\usepackage[colorlinks=true,linkcolor=blue!60!black,citecolor=teal!70!black,
            urlcolor=blue!60!black]{hyperref}
\usepackage{cleveref}

\theoremstyle{plain}

\theoremstyle{definition}

\theoremstyle{remark}

\newcommand{\R}{\mathbb{R}}

\newcommand{\sft}{\operatorname{softmax}}

\DeclareMathOperator*{\argmax}{arg\,max}

\title{%
  \textbf{4-Tensor Attention Model for Semantic Physical Reality}%
}

\author{%
  Jongwook Kim, Sangheon Yun\\
  \small IndigoWave, Center for Quantum Spacetime, Sogang University,\\
  \small 35 Baekbeom-ro, Mapo-gu, Seoul 04107, Republic of Korea\\
  \small \texttt{dr.jongwookkim@gmail.com}
}
\date{\today}

\begin{document}

\maketitle

\begin{abstract}
We describe a 4-tensor attention model that predicts the next semantic state of a scene, for video generation and robot planning. A window of states has positions $(x,t)$ and two fibers, a semantic fiber and a temporal-context fiber, and one softmax normalizes attention jointly over the window. Frames and an agent's situation are written as those states; the encoder, the renderer, and the planner remain outside the update. To test the update on its own, we train on ROCStories, where each window poses the same next-sentence task at the semantic layer. On a one-step prediction of $S(4)$ from $S(0)$ through $S(3)$, the mean paired reduction in cross-entropy over seeds $41$, $42$, and $43$ is $4.9\% \pm 2.7$ at $H{=}2$, $L{=}2$, $2.0\% \pm 0.9$ at $H{=}4$, $L{=}2$, and $4.8\% \pm 5.1$ at $H{=}4$, $L{=}3$, where the second figure is the sample standard deviation. With the window fixed at $T{=}3$, a two-step rollout reduces cross-entropy by $2.8\% \pm 1.1$ on $S(3)$ and by $9.9\% \pm 4.5$ on $S(4)$ at $H{=}2$, $L{=}2$, and by $1.8\% \pm 0.3$ and $16.4\% \pm 0.1$ at $H{=}4$, $L{=}2$ over seeds $41$ and $42$. At $H{=}4$, $L{=}2$ the parameter counts are nearly the same, $172.5$M and $175.9$M. On the same two GPUs those 4-tensor runs finished in $2.0$--$2.4$ hours and the baseline runs in $25.1$--$45.2$ hours; the baseline is trained by free-running decoding, one sequential forward pass per target token.
\end{abstract}

\medskip
\noindent Code is available at \url{https://github.com/indigowave511/semantic-world}.


\tableofcontents
\newpage

\section{Introduction}
\label{sec:intro}

We posit that the semantic representation of a physical scene is a cognitive compression of that scene, and that a world model can therefore advance a scene by evolving its semantic description in discrete steps rather than by interpolating sensory data. This paper describes the update at the center of that program: an attention mechanism that predicts the next semantic state from a window of previous states.

Modeling temporally coherent entities and environments is a central problem both for video generation \cite{brooks2024sora, ho2022imagenvideo} and for world models used in planning and control \cite{hafner2023dreamerv3, brohan2023rt2}. Most current systems operate on continuous sensory representations, in pixel or latent space, and can struggle to keep scenes consistent over long horizons. Part of the difficulty is that semantically relevant state and perceptual detail are mixed in the same dense representation. Latent predictive architectures \cite{lecun2022path, bardes2024vjepa} and discrete visual tokenization with language models \cite{yu2023magvit, kondratyuk2023videopoet} separate the two more explicitly. When such tokens are flattened into a single autoregressive sequence, however, spatially simultaneous content is given an order it does not have, and temporal structure must be recovered from one-dimensional position alone. Video transformers avoid a single raster order by attending jointly over all spatiotemporal tokens, or by factorizing attention into spatial and temporal stages \cite{arnab2021vivit}; each token, however, still carries a single feature vector in which content and temporal context are entangled.

Our approach compresses the content of a scene at a fixed time $t$ into a one-dimensional semantic representation and keeps time as a separate axis. Describing a video as a temporal sequence of sentences has precedent in paragraph captioning \cite{yu2016paragraph}, where the state of earlier sentences conditions the next; there the sentences describe frames that have been observed, whereas here the next semantic state is predicted before it is observed. Each state carries two fibers: a semantic fiber of width $d_x$ and a temporal-context fiber of width $d_t$. A window of states is then a rank-4 tensor, and the update, which we call \emph{4-tensor attention}, acts on that tensor directly rather than on a flattened sequence.

\subsection{Implementation Scheme}

\begin{figure}[htbp]
    \centering
    \begin{tikzpicture}[
        node distance=1.5cm and 2cm,
        data/.style={draw=black!70, fill=blue!5, rectangle, rounded corners,
                     minimum width=3cm, minimum height=1.2cm, align=center, font=\small},
        model/.style={draw=black, fill=gray!10, rectangle, thick,
                      minimum width=3cm, minimum height=1.2cm, align=center, font=\small\bfseries},
        attn/.style={draw=teal!80!black, fill=teal!5, rectangle, rounded corners, thick,
                     minimum width=3cm, minimum height=1.2cm, align=center, font=\small\bfseries},
        arrow/.style={-{Stealth[scale=1.2]}, thick, draw=black!80}
    ]

    \node[data] (video_in) {Video Frames\\(Visual Domain)};
    \node[model] (i2t) [right=of video_in] {Image-to-Text\\Transformation};
    \node[data] (sem_in) [right=of i2t] {Semantic States\\$S(0), S(1), S(2)$};

    \node[attn] (attention) [below=of sem_in] {4-Tensor Attention\\Mechanism};

    \node[data] (sem_out) [below=of attention] {Predicted States\\$\hat{S}(1), \hat{S}(2), \hat{S}(3)$};
    \node[model] (t2i) [left=of sem_out] {Text-to-Image\\Transformation};
    \node[data] (video_out) [left=of t2i] {Predicted Frames};

    \draw[arrow] (video_in) -- (i2t);
    \draw[arrow] (i2t) -- (sem_in);
    \draw[arrow] (sem_in) -- node[right, font=\footnotesize, xshift=0.1cm] {Window of states} (attention);
    \draw[arrow] (attention) -- node[right, font=\footnotesize, xshift=0.1cm] {Next-state prediction} (sem_out);
    \draw[arrow] (sem_out) -- (t2i);
    \draw[arrow] (t2i) -- (video_out);

    \end{tikzpicture}
    \caption{The intended generative pipeline. Frames are compressed into semantic states, the 4-tensor attention update predicts the next states, and the predicted states are rendered back into frames. This paper implements and evaluates only the 4-tensor attention block (teal).}
    \label{fig:pipeline}
\end{figure}
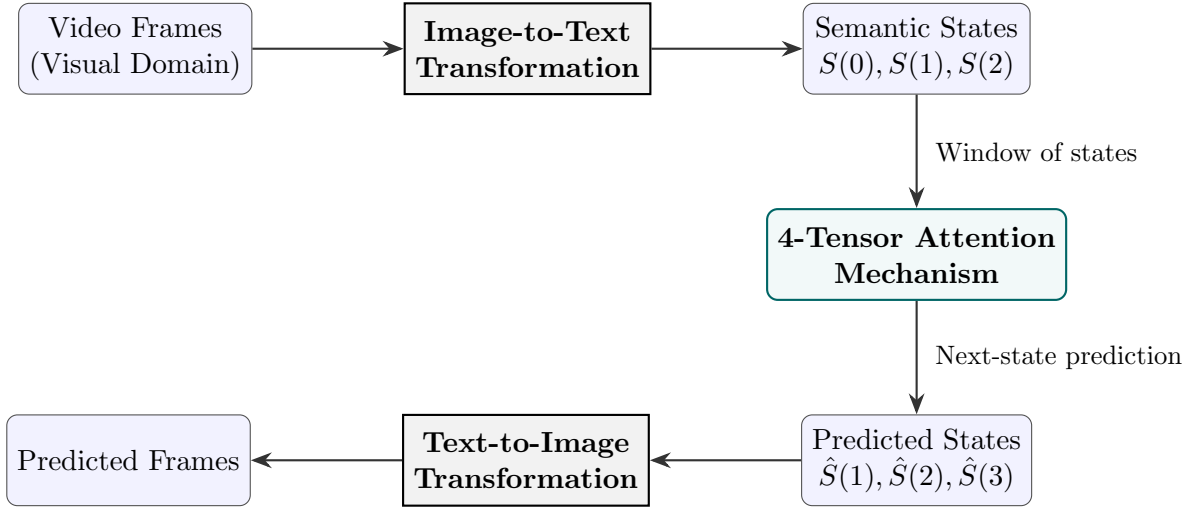

In the intended pipeline, semantic states are extracted from video frames at fixed intervals by an image-to-text model. The resulting sequence of textual states is processed by 4-tensor attention to predict future states, and the predicted states are rendered back into frames by a text-to-image model (\cref{fig:pipeline}). The prediction step is where temporal consistency is decided; the encoder and the renderer determine only how faithfully a state is read from and written back to the visual domain.

\subsection{Scope of This Paper}
\label{sec:scope}

This paper tests one component of that program: whether the 4-tensor update predicts the next semantic state better than a flattened sequence model when both read the same window (\cref{sec:baseline,sec:experiment-data}). The image-to-text encoder, the text-to-image renderer, and any planner are outside this test. To isolate the update from visual encoding and decoding, the experiments operate directly at the semantic layer, on narrative windows from ROCStories, where each window poses the same next-sentence task.

The separation of the two fibers may matter beyond sensory world models, as a representation in which transition rules could be learned apart from the objects they apply to. We return to this question in \cref{sec:discussion}.

\subsection{The Semantic State Sequence}

Let an observation be represented as a sequence of semantic states extracted at regular intervals. To give every state the same length $N$, shorter states are padded with a null token $0$. For an agent waiting for an elevator, with $N=8$:
\begin{quote}
\ttfamily
S(t=0): [I, am, in, front, of, the, elevator, 0]\\
S(t=1): [I, am, looking, at, the, floor, indicator, 0]\\
S(t=2): [The, elevator, arrives, at, my, floor, 0, 0]
\end{quote}
A window of length $T$ collects the most recent $T$ states. For $T=3$,
\begin{equation}
    X(t=2) \leftarrow \{ S(0), S(1), S(2)\}.
\end{equation}
The running example in this section uses $T=3$ and $N=8$. The experiments use $T=4$ and $N=16$ (\cref{sec:corpus}).

\subsection{Rank-4 Representation}

The window is a rank-4 tensor
\begin{equation}
    X \in \R^{N \times d_x \times T \times d_t},
\end{equation}
with the states $S(0), \ldots, S(T{-}1)$ arranged along the temporal axis. Compared with standard scaled dot-product attention, every index is doubled. A sequence position becomes a pair $(x, t)$, where $x$ is the token position within a state and $t$ is the temporal slice. The feature dimension becomes a pair of fibers: the semantic $d_x$-fiber and the temporal-context $d_t$-fiber. Input fiber indices are written $(n, q)$ and output fiber indices $(m, p)$, with $n, m$ on the $d_x$-fiber and $q, p$ on the $d_t$-fiber.

The query, key, and value tensors are joint bilinear contractions over both input fibers,
\begin{equation}
\begin{aligned}
    Q_{xm;tp} &= \sum_{n,q} W^Q_{mn,pq}\, X_{xn;tq}, \\[4pt]
    K_{xm;tp} &= \sum_{n,q} W^K_{mn,pq}\, X_{xn;tq}, \\[4pt]
    V_{xm;tp} &= \sum_{n,q} W^V_{mn,pq}\, X_{xn;tq},
\end{aligned}
\end{equation}
with $W^Q, W^K, W^V \in \R^{d_x \times d_x \times d_t \times d_t}$.

The two fibers enter as distinct tensor factors: each token is embedded as an outer product of a semantic vector and a temporal-context vector (\cref{sec:4tensortransformer}), and the readout ties the semantic fiber to the vocabulary (\cref{sec:forward_pass}). The projection $W_{mn,pq}$, however, is a general bilinear map on the joint $(n,q)$ space and admits arbitrary interaction between the fibers. The architecture therefore does not force transition rules to separate from content. It keeps a representation in which such a separation can be learned and tested.

\subsection{Score Tensor and Joint Softmax}

The score between a query at $(x,t)$ and a key at $(x',t')$ contracts both fibers. To avoid a clash with the state sequence $S(t)$, the unnormalized score tensor is written $E$:
\begin{equation}
    E_{xt,\,x't'} = \frac{1}{\sqrt{d_x d_t}} \sum_{m,p} Q_{xm;tp}\, K_{x'm;t'p}.
\end{equation}
The factor $1/\sqrt{d_x d_t}$ is the standard variance correction for an inner product over $d_x d_t$ coordinates.

The softmax normalizes jointly over the full grid of key positions $(x',t')$, so that a query may draw on any token of any state in the window:
\begin{equation}
    A_{xt,\,x't'} = \frac{\exp(E_{xt,\,x't'})}{\displaystyle\sum_{x'',t''}\exp(E_{xt,\,x''t''})}.
\end{equation}
Apart from padding keys, which are masked (\cref{sec:steps25}), no mask is applied. In particular there is no temporal causal mask: every query attends to every slice of the window.

\subsection{Attention Output}

The attention output is the $A$-weighted sum of values over the key grid,
\begin{equation}
    y_{xm;tp} = \sum_{x',t'} A_{xt,\,x't'}\, V_{x'm;t'p},
\end{equation}
and has the same shape $\R^{N \times d_x \times T \times d_t}$ as the input $X$.

\subsection{Prediction Targets and the Learning Objective}

In the full model the attention block is followed by a residual connection, a position-wise MLP, and a final LayerNorm, and the result is read out to vocabulary logits at every position $(x, t)$ (\cref{sec:forward_pass,sec:pipeline}). Write $F$ for this map from the window to the logits. The readout at slice $t$ is a prediction of the next state,
\begin{equation}
    p_{xv}(t) = \sft_v\big(F(X)_{x,t,v}\big)
    \quad\text{predicts}\quad S(t{+}1)_x .
\end{equation}
An input window $S(0), S(1), S(2)$ therefore yields predictions $\hat{S}(1), \hat{S}(2), \hat{S}(3)$, and the training targets are the window shifted by one slice:
\begin{quote}
\ttfamily
S(t=1): [I, am, looking, at, the, floor, indicator, 0]\\
S(t=2): [The, elevator, arrives, at, my, floor, 0, 0]\\
S(t=3): [The, elevator, doors, open, in, front, of, me]
\end{quote}
Training minimizes the cross-entropy of all $T$ predicted states against their targets, with padding positions ignored.

Because the joint softmax carries no temporal causal mask, for $t = 0, \ldots, T{-}2$ the target $S(t{+}1)$ of slice $t$ is present in input slice $t{+}1$. The terms $S(1), \ldots, S(T{-}1)$ of the training loss therefore read sentences already in the input. Prediction of a sentence absent from the input is measured only at the last slice:
\begin{equation}
    \text{Evaluation metric} \leftarrow \mathrm{CE}\big(\hat{S}(T) \,\big\|\, S(T)\big).
\end{equation}
In the running example this is $\mathrm{CE}(\hat{S}(3) \,\|\, S(3))$, and in the experiments $\mathrm{CE}(\hat{S}(4) \,\|\, S(4))$. All tokens of $\hat{S}(T)$ are predicted in one parallel forward pass, and the tokens of $S(T)$ enter only as targets.

\subsection{Autoregressive Inference}

The generative task is to continue the sequence beyond the window:
\begin{quote}
\ttfamily
S(t=3): [The, elevator, doors, open, in, front, of, me]\\
S(t=4): [I, step, inside, the, elevator, 0, 0, 0]\\
S(t=5): [The, doors, close, behind, me, 0, 0, 0]\\
S(t=6): [I, press, the, button, for, my, floor, 0]
\end{quote}
The predicted state $\hat{S}(3)$ is first converted to tokens, $\hat{S}(3)_x = \argmax_v p_{xv}(2)$, or by sampling. The window then slides by one state,
\begin{equation}
    X' = \{S(1), S(2), \hat{S}(3)\} \in \R^{N \times d_x \times T \times d_t},
\end{equation}
where the three states again occupy slices $t = 0, 1, 2$ and are embedded by the same input construction. The same weights are applied, and the next state is read from the last slice:
\begin{equation}
    \hat{S}(4)_x = \argmax_v\, F(X')_{x,\,T-1,\,v}.
\end{equation}
Repeating the step with $X'' = \{S(2), \hat{S}(3), \hat{S}(4)\}$ gives $\hat{S}(5)$, and so on. Ground-truth states from $S(3)$ onward never enter the window; each step conditions on the model's own earlier predictions. The window can thus be advanced repeatedly with fixed weights. The tables in \cref{sec:experiment-data} report the single step $\mathrm{CE}(\hat{S}(4) \,\|\, S(4))$. The two-step rule, on a window of length $3$, is \cref{sec:protocol-rollout}.

\section{4-Tensor Transformer Design}
\label{sec:4tensortransformer}

\subsection{Overview of the Tensor Pipeline}

The 4-Tensor Transformer takes as its base unit the state tensor at a given temporal slice $t$, defined as: 
\begin{equation}
  S_{xn;q}(t) \in \mathbb{R}^{N \times d_x \times d_t}
  \label{eq:rank3_input}
\end{equation}

The rank-4 input tensor $X_{xn;tq}$ is constructed via a sequential pipeline rather than a simultaneous summation. Each stage injects a single piece of information into a specific fiber:
\begin{equation}
    S(t)
    \;\xrightarrow{\;+\,P_{xn}\;}\;
    \tilde{S}(t)
    \;\xrightarrow{\;\text{stack over }t\;}\;
    X^{(0)}_{xn;tq}
    \;\xrightarrow{\;+\,T^{\mathrm{emb}}_{xtq}\;}\;
    X_{xn;tq}
\end{equation}

Here, $P_{xn}$ is a spatial position embedding from $\mathbb{R}^N$ into the semantic $d_x$-fiber, and $T^{\mathrm{emb}}_{xtq}$ is a temporal position embedding added from $\mathbb{R}^T$ into the temporal $d_t$-fiber. This ordering ensures that the spatial token position ($x$) and temporal sequence position ($t$) remain analytically separable within the unified rank-4 object.

\subsection{Stage-by-Stage Logic}

\subsubsection*{Stage A: Initialization by Dual-Dictionary Token Embedding}
Traditionally, a token is embedded via a single dictionary lookup into a semantic vector space. However, because the temporal-context $d_t$-fiber operates as a first-class axis in our rank-4 formulation, the token's initial embedding must explicitly bifurcate into two independent dictionaries:
\begin{equation}
    E_{\text{sem}} : V \to \mathbb{R}^{d_x}, \qquad E_{\text{temp}} : V \to \mathbb{R}^{d_t}
\end{equation}

The token embedding is rank-1 across the two fibers: the base state $S(t)$ at position $x$ is the outer product of the two dictionary vectors. For each discrete temporal slot $t \in \{0,\dots,T-1\}$, a token at spatial position $x \in \{0,\dots,N-1\}$ is embedded via two separate dictionaries:
\begin{equation}
    S_{xn;q}(t) = E_{\text{sem}}(\text{tok}_{x,t})_n \cdot E_{\text{temp}}(\text{tok}_{x,t})_q \;\in\; \mathbb{R}^{N \times d_x \times d_t}
\end{equation}
Thus, $S(t)$ is rank-3 from the outset, intrinsically carrying content along both the semantic $d_x$-fiber and the temporal-context $d_t$-fiber prior to any positional injection.

\subsubsection*{Stage B: Spatial Position Injection}
The spatial position embedding $P_{xn} \in \mathbb{R}^{N \times d_x}$ is added independently per temporal slice:
\begin{equation}
    \tilde{S}_{xn;q}(t) = S_{xn;q}(t) + P_{xn}
\end{equation}
Because spatial placement governs static semantic identity, $P_{xn}$ resides entirely within the $d_x$-fiber. It broadcasts identically across the $d_t$-fiber ($q$), preserving the rank-3 structure.

\subsubsection*{Stage C: Temporal Stacking}
The $T$ sentence tensors are stacked along a newly initialized temporal axis:
\begin{equation}
    X^{(0)}_{xn;tq} = \tilde{S}_{xn;q}(t)
\end{equation}
This operation maps the sequence into the rank-4 configuration $(N, d_x, T, d_t)$. Because $P_{xn}$ operates independently per sentence and broadcasts over $q$, the operations of adding $P$ and stacking over $t$ are mathematically commutative.

\subsubsection*{Stage D: Temporal-Order Injection}
Finally, the temporal-order embedding $T^{\mathrm{emb}}_{xtq} \in \mathbb{R}^{N \times T \times d_t}$ is added, broadcasting strictly over the semantic $d_x$-fiber ($n$):
\begin{equation}
    X_{xn;tq} = X^{(0)}_{xn;tq} + T^{\mathrm{emb}}_{xtq}
\end{equation}
While Stage A establishes the token-level context, Stage D establishes the rigorous positional trajectory across the causal surface.

\subsection{Joint-Phase Temporal Embedding}

Crucially, the temporal embedding $T^{\mathrm{emb}}$ must depend on both the temporal slice $t$ and the spatial column $x$. A naive parameterization $T^{\mathrm{emb}}_{tq}$ would assign the identical temporal signal to every spatial token within a given slice $t$. In scenarios where the same entity appears at different times (e.g., the token \texttt{[I]} appearing at $t=0$ and $t=4$ in our physical elevator progression), a column-independent signal fails to distinguish the token's shifting causal roles.

To resolve this, we formulate $T^{\mathrm{emb}}_{xtq}$ as a joint-phase embedding:
\begin{equation}
T^{\mathrm{emb}}_{xtq} =
\begin{cases}
\sin\!\left(\dfrac{x+t}{10000^{2i/d_t}}\right), & q = 2i \\[10pt]
\cos\!\left(\dfrac{x+t}{10000^{2i/d_t}}\right), & q = 2i+1
\end{cases}
\qquad i = 0,\dots,\lfloor d_t/2 \rfloor - 1
\end{equation}

This formulation satisfies two properties of the phase $x+t$:
\begin{enumerate}
    \item \textbf{Column Independence:} For a fixed $t$, $T^{\mathrm{emb}}_{xtq} \neq T^{\mathrm{emb}}_{x'tq}$ whenever $x \neq x'$. The same phase gives $T^{\mathrm{emb}}_{x,t,q} = T^{\mathrm{emb}}_{x-1,t+1,q}$. Positions that share a value of $x+t$ are distinguished once $T^{\mathrm{emb}}$ is added to the spatial embedding $P_{xn}$.
    \item \textbf{Offset Consistency:} For a fixed shift $k$, the phase increment depends only on $k$ and the frequency index $i$, and a linear map $M_k$ on the $q$-fiber, independent of $x$, carries $T^{\mathrm{emb}}_{x,t,\cdot}$ to $T^{\mathrm{emb}}_{x,t+k,\cdot}$. The implemented slide does not use $M_k$. Each new window is written again onto positions $t \in \{0,\ldots,T-1\}$, and $T^{\mathrm{emb}}$ is recomputed from those positions.
\end{enumerate}

\subsection{Final Assembly of the Input Tensor}

The complete rank-4 input tensor is thus formalized through the direct superposition of these structured stages:
\begin{equation}
    \boxed{ X_{xn;tq} = \underbrace{\left[ S_{xn;q}(t) + P_{xn} \right]}_{\text{Stages A \& B (Per-slice)}} + \underbrace{T^{\mathrm{emb}}_{xtq}}_{\text{Stage D (Joint-phase)}} }
\end{equation}
yielding the exact causal dimensionality required:
\begin{equation}
    X \in \mathbb{R}^{N \times d_x \times T \times d_t}
\end{equation}

This boxed formulation is the input to the 4-Tensor Attention block. The sum $S + P + T^{\mathrm{emb}}$ is one tensor; the two fibers stay distinct factors, and $W_{mn,pq}$ may mix them.

\section{Forward Pass and Pipeline Overview}
\label{sec:forward_pass}

Following the construction of the completed rank-4 input tensor $X_{xn;tq}$, this section covers the remainder of a single-layer, single-head forward pass through to the training loss. Indices are zero-based, $x \in \{0,\ldots,N-1\}$ and $t \in \{0,\ldots,T-1\}$, as in \cref{sec:storage}. The pipeline encompasses the attention operation, residual connections with dropout, the position-wise MLP block, readout-boundary LayerNorm, vocabulary readout, and the final cross-entropy loss, including the $t$-index shift that connects the attention update to the prediction targets.

\subsection{Pipeline Overview}

The full transformation from the input tensor to the scalar loss $\mathcal{L}$ is strictly sequentially ordered:

\begin{equation}
\label{eq:pipeline}
\begin{aligned}
X_{xn;tq} \;\xrightarrow{\text{4-Tensor Attention}}\; y_{xm;tp} \;\xrightarrow{+\,X,\,\text{drop}}\; X'_{xn;tq} \;\xrightarrow{\text{MLP}}\; &Z_{xn;tq} \\
&\;\;\Big\downarrow \scriptstyle{+\,X',\,\text{drop}} \\
\mathcal{L} \;\xleftarrow{\text{CE}}\; p_{xv}(t) \;\xleftarrow{\sft_v}\; L_{xv}(t) \;\xleftarrow{\text{readout}}\; X^{\text{norm}}_{xn;tq} \;\xleftarrow{\text{LN}_{nq}}\; &\hat{X}_{xn;tq}
\end{aligned}
\end{equation}

Where the intermediate variables are defined as follows:
\begin{itemize}
    \item $y_{xm;tp}$: The raw rank-4 output of the attention mechanism.
    \item $X'_{xn;tq}$: The intermediate state following the first residual connection and dropout.
    \item $Z_{xn;tq}$: The unflattened output of the position-wise MLP block.
    \item $\hat{X}_{xn;tq}$: The completed rank-4 representation for the layer following the second residual connection and dropout.
    \item $X^{\text{norm}}_{xn;tq}$: The state after applying LayerNorm over the semantic and temporal-context fibers ($n, q$), preparing the tensor for vocabulary readout.
\end{itemize}

The construction of $y_{xm;tp}$ (including the QKV generation, score tensor computation, and joint softmax) initiates the stage. The subsequent sections detail the pipeline beginning directly from the first residual connection.

\subsection{Residual After Attention}

Following the attention mechanism, the first residual connection is applied:
\begin{equation}
    X'_{xn;tq} = X_{xn;tq} + \mathrm{dropout}(y_{xn;tq})
\end{equation}

The indices align directly: the attention output $y$ resides in the exact same $(x, n, t, q)$ space as the input $X$. This dimensional consistency is guaranteed because the query, key, and value tensors were all constructed by contracting the identical $(n,q)$ fibers via the weight matrices $W_Q, W_K, W_V \in \R^{d_x \times d_x \times d_t \times d_t}$. Dropout is applied exclusively to the residual branch to preserve the main identity pathway.

\subsection{Event Position MLP Block}

The feed-forward network operates independently across each $(x, t)$ event position on the causal surface. Because the semantic ($n$) and temporal-context ($q$) fibers represent distinct algebraic spaces, we impose no separability assumption between them at this stage. Instead, the $(n, q)$ slice at each fixed $(x, t)$ is jointly flattened into a single feature vector of dimension $d_x d_t$ before applying the non-linear transformations:
\begin{equation}
    Z_{xn;tq} = \mathrm{unflatten}\Big( W_2\,\sigma\big(W_1\, \mathrm{flatten}(X'_{x\cdot;t\cdot}) + b_1\big) + b_2 \Big)
\end{equation}
where $\mathrm{flatten}: \R^{d_x \times d_t} \to \R^{d_x d_t}$ maps the matrix slice into a vector, and $\mathrm{unflatten}$ restores the original rank-4 structure.

The projection weights are defined as $W_1 \in \R^{d_{ff} \times (d_x d_t)}$ and $W_2 \in \R^{(d_x d_t) \times d_{ff}}$, with $\sigma$ denoting the GELU activation function. To maintain the standard transformer capacity scaling relative to the joint input dimension, the hidden layer width is explicitly set to the standard expansion ratio: $d_{ff} = 4 d_x d_t$.

\subsection{Residual After MLP and Readout-Boundary Normalization}

Following the MLP block, the second residual connection is applied to produce the completed layer output:
\begin{equation}
    \hat{X}_{xn;tq} = X'_{xn;tq} + \mathrm{dropout}(Z_{xn;tq})
\end{equation}

The residual stream itself remains the raw sum of the block input and the dropped branch outputs. In the reduced-to-practice stack of \cref{sec:pipeline}, each branch input is LayerNorm'd before attention and before the MLP; those normalizations are not placed after the residual add. The residual stream can therefore still grow across layers. If left unscaled at the readout, that magnitude saturates the vocabulary softmax and collapses the cross-entropy gradient. The readout-boundary LayerNorm below is retained in addition to the per-branch normalizations.

To stabilize the output logits without interfering with the internal attention dynamics, we introduce a Layer Normalization strictly at the readout boundary of the final layer. Consistent with the MLP block, this normalization operates over the jointly flattened semantic and temporal-context fibers, imposing no separability assumption. For the final layer's output $\hat{X}_{xn;tq}$, the normalized tensor $X^{\text{norm}}_{xn;tq}$ is computed as:
\begin{equation}
    X^{\text{norm}}_{x\cdot;t\cdot} = \mathrm{unflatten}\Big( \mathrm{LayerNorm}_{d_x d_t}\big(\mathrm{flatten}(\hat{X}_{x\cdot;t\cdot})\big) \Big)
\end{equation}
The normalization is applied independently at each spatial-temporal position $(x, t)$ over the $d_x d_t$ dimensional space, incorporating standard learnable affine parameters $\gamma, \beta \in \R^{d_x d_t}$. By operating locally at each $(x, t)$ coordinate, the normalization strictly preserves the independence of causal slots and prevents any temporal information leakage.

\subsection{Vocabulary Readout and Structured Factorization}

To project the normalized state $X^{\text{norm}}_{xn;tq}$ into the vocabulary, the general linear map is a dense tensor $W_{\text{out}} \in \R^{d_x \times d_t \times |V|}$. The readout used here factorizes each vocabulary column as a rank-1 tensor on $(n, q)$.

The semantic factor is the token embedding $E_{\text{sem}} \in \R^{|V| \times d_x}$, tied to Stage~A, and it contributes no new parameters. The temporal-context factor is a learned matrix $W' \in \R^{d_t \times |V|}$. Each vocabulary item is therefore one rank-1 pair $(n,q)$, the same boundary described as a rank-1 readout functional in \cref{sec:discussion}.

The factorization is:
\begin{equation}
    W_{\text{out},\,nq;v} = E_{\text{sem}}[v,n] \cdot W'_{q,v}
\end{equation}
where $v$ indexes the vocabulary. Substituting this into the logit computation over the flattened $(n, q)$ slice yields:
\begin{equation}
    L_{xv}(t) = \sum_{n,q} X^{\text{norm}}_{xn;tq}\, E_{\text{sem}}[v,n]\, W'_{q,v}
\end{equation}

\subsubsection{Sequence-Independent Matrix Pre-computation}

To evaluate this logit efficiently and avoid materializing a prohibitively large sequence-dependent intermediate tensor of size $\mathcal{O}(NT d_t |V|)$, we define an optimal evaluation order. We pre-compute a strictly sequence-independent projection matrix $M$:
\begin{equation}
    M_{\phi(n,q),\,v} = E_{\text{sem}}[v,n]\, W'_{q,v}, \qquad M \in \R^{d_x d_t \times |V|}
\end{equation}
where $\phi(n,q) = n d_t + q$ is the identical joint-flattening bijection used in the MLP block. The final vocabulary logits $L_{xv}(t)$ are then computed via a direct matrix multiplication between the jointly flattened representation $X^{\text{norm}}$ and the factorized matrix $M$:
\begin{equation}
    L_{xv}(t) = \sum_{n,q} X^{\text{norm}}_{xn;tq}\, M_{\phi(n,q),\,v}
\end{equation}
By decoupling the lexical projection from the sequence length, this factorized construction strictly preserves both the structural meaning of the distinct fibers and the computational tractability of the forward pass.

\subsection{Structural Constraints Imposed by the Loss Objective}

The mathematical structure of the vocabulary readout is not arbitrary; it is strictly dictated by the terminal loss function. Because the final training objective evaluates the predicted semantic states $\hat{S}(t)$ against the ground truth $S(t)$ pointwise across both the sequence position $x$ and the temporal step $t$, these two indices must be explicitly preserved. Conversely, the semantic ($n$) and temporal-context ($q$) fibers represent internal latent coordinates that do not exist in the discrete target space. Consequently, they must be contracted away.

This deterministic survival of indices---preserving $(x, t)$ while eliminating $(n, q)$---requires a mapping from the normalized latent state $X^{\text{norm}}_{xn;tq}$ to the vocabulary logits $L_{xv}(t)$. The most general linear transformation is a joint bilinear contraction over the $(n, q)$ plane:
\begin{equation}
    L_{xv}(t) = \sum_{n,q} X^{\text{norm}}_{xn;tq} \, W_{\text{out},\,nq;v}
\end{equation}
where the readout tensor is defined as $W_{\text{out}} \in \R^{d_x \times d_t \times |V|}$. 

That dense contraction is the general linear map. The implemented readout replaces it by the rank-1 factorization $W_{\text{out},\,nq;v} = E_{\text{sem}}[v,n]\, W'_{q,v}$, one rank-1 pair $(n,q)$ for each vocabulary item.

\subsection{Mapping to the Vocabulary Simplex}

The raw logits $L_{xv}(t)$ derived from the readout projection are mapped onto the vocabulary simplex via a standard softmax operation over the vocabulary axis $v$:
\begin{equation}
    p_{xv}(t) = \frac{\exp(L_{xv}(t))}{\sum_{v'} \exp(L_{xv'}(t))}
\end{equation}

This transformation is computed independently at each spatial-temporal coordinate $(x,t)$, yielding a discrete probability distribution $p_{xv}(t) \in \Delta^{|V|-1}$ over the vocabulary space. This tensor represents the model's predicted likelihood for each discrete semantic token $v$ at position $x$ and time step $t$. 

While the explicit materialization of $p_{xv}(t)$ is strictly required for inference algorithms (e.g., autoregressive sampling or argmax decoding), during training, this probability mapping is analytically absorbed directly into the cross-entropy objective to guarantee numerical stability.

\subsection{From Embedding-Level Loss to Cross-Entropy Loss}

While traditional continuous generative architectures often formulate training objectives as direct regression losses at the embedding level (e.g., minimizing $\|\hat{X}(t) - X(t)\|^2$), our discrete semantic framework evaluates predictions directly over the vocabulary simplex. 

Transitioning from continuous vector space interpolation to discrete semantic evolution, the training objective is formulated as the cross-entropy loss over the predicted vocabulary distribution:
\begin{equation}
    \mathcal{L} = -\sum_{t=0}^{T-1} \sum_{x=0}^{N-1} \sum_{v \in \mathcal{V}} y_{xv}(t) \log p_{xv}(t)
\end{equation}
where $x \in \{0,\ldots,N-1\}$ and the logit slice $t \in \{0,\ldots,T-1\}$ is the distribution for sentence $S(t{+}1)$. Here $y_{xv}(t)$ is the target distribution at that position, and $p_{xv}(t)$ is the softmax of the readout logits. This formulation bridges the continuous hidden representations of the causal surface with the discrete semantic vocabulary space, completing the end-to-end differentiable training pipeline.

\section{Enabling Construction of the Forward Pipeline}
\label{sec:pipeline}

This section specifies the forward pipeline that was reduced to practice. The algebraic maps of the preceding sections are unchanged. What is fixed here is the order of operations, the in-memory axis order, the numerical guards, and which tensors are parameters. An implementation that preserves every shape in \cref{tab:storage} can still compute a different function if any check in \cref{sec:faithfulness} is omitted. Storage indices below are zero-based.

\subsection{Index convention and storage}
\label{sec:storage}

Theory indices remain $X_{xn;tq}$ and $W_{mn,pq}$. The stored activation places the position axes adjacent and the fibers last, so that the flattening
\begin{equation}
    \phi(n,q) = n\, d_t + q
\end{equation}
is a reshape with no permutation. The same $\phi$ is used for the attention contraction, the MLP, the readout-boundary LayerNorm, and the vocabulary matrix $M$. A tensor whose last two axes are $(d_t, d_x)$ has the same product $d_x d_t$ and would pass a product check while applying the wrong $\phi$. Fiber checks therefore compare the ordered pair $(d_x, d_t)$.

\begin{table}[t]
\centering
\small
\caption{Storage layout of the reduced-to-practice pipeline. The batch axis $B$ is present on every public tensor, including the case $B=1$.}
\label{tab:storage}
\begin{tabular}{@{}llp{0.42\textwidth}@{}}
\toprule
Object & Shape & Index order \\
\midrule
\texttt{token\_ids}, $Y$ & $(B, T, N)$ & $[b, t, x]$ \\
$X, Q, K, V, \mathrm{Attn}$ & $(B, N, T, d_x, d_t)$ & $[b, x, t, n, q]$ \\
\texttt{pad\_mask} & $(B, N, T)$ & $[b, x, t]$, true on a PAD key \\
$W^Q, W^K, W^V$ & $(d_x, d_t, d_x, d_t)$ & $[m, p, n, q]$ \\
scores, attention weights & $(B, N, T, N, T)$ & $[b, x, t, x', t']$ \\
logits & $(B, N, T, |V|)$ & $[b, x, t, v]$ \\
\bottomrule
\end{tabular}
\end{table}

The weight tensor is the view of a matrix,
\begin{equation}
    W[m, p, n, q]
    = \mathrm{mat}\big[m\, d_t + p,\; n\, d_t + q\big],
\end{equation}
so $W.\mathrm{view}(d_x d_t,\, d_x d_t)$ is the linear map $(n,q)\mapsto(m,p)$ with no copy. Logits are stored $(B, N, T, |V|)$ and targets are stored $(B, T, N)$. The bridge is the permutation $(b, t, x)\mapsto(b, x, t)$. When $N=T$ that permutation preserves the shape. A missing permutation is therefore caught by checking that the target has shape $(B, T, N)$ before the permutation is applied.

A stored example supplied to training has shape $(B, T{+}1, N)$, the sentences $S(0),\ldots,S(T)$. The forward input is the prefix $S(0),\ldots,S(T{-}1)$, of shape $(B, T, N)$. An array of rank $2$ and shape $(T{+}1, N)$ is rejected. The caller inserts the batch axis.

\subsection{Parameters, buffers, and initialization}

The learned parameters are the following.
\begin{itemize}
    \item $E_{\mathrm{sem}}\in\R^{|V|\times d_x}$ and $E_{\mathrm{temp}}\in\R^{|V|\times d_t}$, with the row at the padding index held at zero and excluded from the gradient.
    \item For each layer and each head, three maps $W^Q, W^K, W^V$, each stored as above. Xavier uniform initialization is applied to the flat $(d_x d_t)\times(d_x d_t)$ matrix and then viewed into $(m,p,n,q)$. Initializing the four-dimensional array directly mis-estimates the fan-in.
    \item For each layer, two LayerNorm affine pairs of width $d_x d_t$ (one before attention, one before the MLP), and one MLP of width $d_{ff}=4\, d_x d_t$.
    \item One further LayerNorm of width $d_x d_t$ after the last layer.
    \item The readout factor $W'\in\R^{d_t\times|V|}$, Xavier-uniform initialized.
\end{itemize}
The spatial table $P_{xn}$ and the joint-phase table $T^{\mathrm{emb}}_{xtq}$ are fixed sinusoids, recomputed or registered as non-learned buffers. Dropout probability $p$ is a hyperparameter, set to $0.1$ in the reported runs. Evaluation disables dropout.

\subsection{Step 1: construction of the input tensor}
\label{sec:step1}

The input $X$ is assembled in four stages. Each stage writes one signal into one fiber.

\paragraph{Stage A, content.}
For a token identifier $\mathrm{tok}_{b,t,x}$,
\begin{equation}
    S_{b,x,n,t,q}
    = E_{\mathrm{sem}}[\mathrm{tok}_{b,t,x}]_n
      \cdot
      E_{\mathrm{temp}}[\mathrm{tok}_{b,t,x}]_q .
\end{equation}
The slice at a padding identifier is the zero tensor, because both embedding rows are zero. No position has been added.

\paragraph{Stage B, spatial position.}
$P\in\R^{N\times d_x}$ depends only on the column $x$. With $i=0,\ldots,\lfloor d_x/2\rfloor-1$,
\begin{equation}
    P_{x,2i}
    = \sin\!\Big(\frac{x}{10000^{2i/d_x}}\Big),
    \qquad
    P_{x,2i+1}
    = \cos\!\Big(\frac{x}{10000^{2i/d_x}}\Big).
\end{equation}
If $d_x$ is odd, the last coordinate remains $0$. $P_{xn}$ is added at every batch index, every sentence index, and every $q$:
\begin{equation}
    \tilde S_{b,x,n,t,q} = S_{b,x,n,t,q} + P_{x,n}.
\end{equation}
Adding $P$ before or after the stack over $t$ is the same map, because $P$ carries no $t$ index.

\paragraph{Stage C, stack.}
The $T$ sentences are already arranged along the sentence axis of the stored tensor $(B, N, T, d_x, d_t)$. Stacking performs no arithmetic.

\paragraph{Stage D, joint-phase temporal embedding.}
$T^{\mathrm{emb}}\in\R^{N\times T\times d_t}$ depends on the sum of the column and the sentence index. With $i=0,\ldots,\lfloor d_t/2\rfloor-1$,
\begin{equation}
    T^{\mathrm{emb}}_{x,t,2i}
    = \sin\!\Big(\frac{x+t}{10000^{2i/d_t}}\Big),
    \qquad
    T^{\mathrm{emb}}_{x,t,2i+1}
    = \cos\!\Big(\frac{x+t}{10000^{2i/d_t}}\Big),
\end{equation}
and the last coordinate remains $0$ when $d_t$ is odd. It is added at every batch index and every semantic coordinate $n$:
\begin{equation}
\label{eq:step1}
    X_{b,x,n,t,q}
    = \tilde S_{b,x,n,t,q} + T^{\mathrm{emb}}_{x,t,q}.
\end{equation}
The phase $x+t$ makes the signal differ across columns of one sentence, and a shift of the sentence index is a linear map on the $q$-fiber. When the window slides, the positions $t \in \{0,\ldots,T-1\}$ are reused and $T^{\mathrm{emb}}$ is rebuilt on them. The maps $W^Q, W^K, W^V$ apply to that re-indexed window.

On the stored layout the broadcasts are $P$ expanded in the sentence and $d_t$ slots, and $T^{\mathrm{emb}}$ expanded in the $d_x$ slot. The completed tensor has shape $(B, N, T, d_x, d_t)$.

A padding slot is not the zero tensor after Step~1. Stages B and D add $P$ and $T^{\mathrm{emb}}$ regardless of the token:
\begin{equation}
    X_{\mathrm{pad}} = P[x] + T^{\mathrm{emb}}[x,t] \neq 0.
\end{equation}
The keys and values built from that slot are nonzero. Which slots are padding is read from the integer token identifiers, not from the norm of $X$.

\subsection{Steps 2--5: one attention head}
\label{sec:steps25}

One head owns three full-fiber maps. The contraction, written on storage indices, is
\begin{equation}
    Q[b,x,t,m,p]
    = \sum_{n=0}^{d_x-1}\sum_{q=0}^{d_t-1}
      W^Q[m,p,n,q]\, X[b,x,t,n,q],
\end{equation}
and likewise for $K$ and $V$ with $W^K$ and $W^V$. Each of $Q, K, V$ has shape $(B, N, T, d_x, d_t)$. In the reduced-to-practice layer this contraction receives $\mathrm{LN}(X)$, as specified in \cref{sec:preln}, rather than the raw residual.

The score is the scaled Frobenius product over the $(m,p)$ slice, jointly over the key grid $(x',t')$:
\begin{equation}
    E[b,x,t,x',t']
    = \frac{1}{\sqrt{d_x d_t}}
      \sum_{m,p}
      Q[b,x,t,m,p]\, K[b,x',t',m,p].
\end{equation}
With $N$ adjacent to $T$, the reshape $(B, N T, d_x d_t)$ followed by a batched matrix product realizes this sum. The score tensor has shape $(B, N, T, N, T)$.

The padding mask is built once per forward call:
\begin{equation}
    \texttt{pad\_mask}[b,x,t]
    = \big(\texttt{token\_ids}[b,t,x] = \texttt{padding\_idx}\big),
\end{equation}
which is the permutation $(b,t,x)\mapsto(b,x,t)$ of a boolean tensor. It is reused at every depth of that call, because later layers rewrite values of $X$ and do not rewrite which slot held which identifier. It is rebuilt on the next call. A generation step that writes a token into a former padding slot must observe the new mask.

The mask is applied on the key axes only, by writing the smallest finite value of the score dtype into $E[b,x,t,x',t']$ whenever $\texttt{pad\_mask}[b,x',t']$ is true. Query axes are left unchanged: a padding query is removed from the loss by the ignore index, and the MLP has no path from one slot to another. The fill value is the finite minimum, not $-\infty$. If every key of a query is padding, a fill of $-\infty$ makes the softmax maximum undefined. The finite minimum leaves a finite maximum, and the softmax returns the uniform distribution on the $NT$ keys.

Joint softmax then normalizes the last two axes together. In exact arithmetic,
\begin{equation}
    A[b,x,t,x',t']
    = \frac{\exp\big(E[b,x,t,x',t']\big)}
           {\sum_{x'',t''}\exp\big(E[b,x,t,x'',t'']\big)}.
\end{equation}
The implementation subtracts the maximum over $(x'',t'')$ before the exponential, which is the stabilization inside a softmax on the flattened key axis of length $NT$. Dropout is not applied to $A$.

The head output contracts $A$ with $V$ and returns to the fiber layout:
\begin{equation}
    \mathrm{Attn}[b,x,t,m,p]
    = \sum_{x',t'}
      A[b,x,t,x',t']\, V[b,x',t',m,p].
\end{equation}
The result has shape $(B, N, T, d_x, d_t)$, identical to the residual stream.

\subsection{Multi-head sum and the Pre-LN layer}
\label{sec:preln}

A head count $H\ge 1$ instantiates $H$ independent copies of Steps~2--5. Each copy keeps fibers of size $d_x$ and $d_t$. The copies are summed, with no factor $1/\sqrt{H}$:
\begin{equation}
    \mathrm{Attn}(U) = \sum_{h=1}^{H} \mathrm{Attn}^{(h)}(U).
\end{equation}
The case $H=1$ is exactly one head. The argument $U$ is a normalized copy of the residual, not the residual itself.

Let $\mathrm{LN}_{d}$ denote LayerNorm over a last axis of length $d=d_x d_t$, with $\varepsilon=10^{-5}$ and learned affine scale and shift, applied independently at each $(b,x,t)$:
\begin{equation}
    \mathrm{LN}_{d}(v)
    = \gamma\odot\frac{v-\mu(v)}{\sqrt{\sigma^2(v)+\varepsilon}}+\beta,
    \qquad
    \gamma,\beta\in\R^{d}.
\end{equation}
The flatten and unflatten use $\phi$. One layer then executes
\begin{align}
    U &= \mathrm{unflatten}\big(\mathrm{LN}^{(1)}_{d}(\mathrm{flatten}(X))\big), \\
    \mathrm{Attn} &= \textstyle\sum_{h}\mathrm{Attn}^{(h)}(U), \\
    X^{\mathrm{attn}} &= X + \mathrm{Dropout}_{p}(\mathrm{Attn}), \\
    U' &= \mathrm{unflatten}\big(\mathrm{LN}^{(2)}_{d}(\mathrm{flatten}(X^{\mathrm{attn}}))\big), \\
    Z &= \mathrm{MLP}(U'), \\
    X^{\mathrm{out}} &= X^{\mathrm{attn}} + \mathrm{Dropout}_{p}(Z).
\end{align}
$\mathrm{LN}^{(1)}$ and $\mathrm{LN}^{(2)}$ are distinct modules. The residual additions use the unnormalized stream $X$ and $X^{\mathrm{attn}}$. The two dropout calls share the rate $p$ and draw independent masks. They are the only dropout sites: embeddings, the pre-layer norms, the joint softmax, the final norm, and the readout are undropped.

The MLP acts at each $(b,x,t)$ on the flattened fiber $v\in\R^{d_x d_t}$:
\begin{equation}
    Z = \mathrm{unflatten}\Big(
        W_2\, \mathrm{GELU}(W_1 v + b_1) + b_2
    \Big),
\end{equation}
with $W_1\in\R^{d_{ff}\times d_x d_t}$, $W_2\in\R^{d_x d_t\times d_{ff}}$, $d_{ff}=4\, d_x d_t$, and
\begin{equation}
    \mathrm{GELU}(u)
    = \tfrac{1}{2}\, u\,
      \Big(1+\mathrm{erf}\big(u/\sqrt{2}\big)\Big).
\end{equation}
The linear maps use the reference implementation's default initialization (Kaiming uniform on the weights, uniform biases of bound $1/\sqrt{d_{\mathrm{in}}}$). The hidden width $4\, d_x d_t$ is the ordinary expansion of a model dimension $d_x d_t$.

$L$ such layers are applied in sequence to the same $X$, each consuming the shared \texttt{pad\_mask} of the call. After the last layer, one further $\mathrm{LN}_{d}$, denoted $\mathrm{final\_norm}$, produces $X^{\mathrm{norm}}$. This norm is required in addition to the per-branch norms, because the residual stream that reaches it is the raw sum.

\subsection{Readout and the training loss}
\label{sec:readout-impl}

The readout does not allocate a parameter tensor of shape $(d_x, d_t, |V|)$. It ties the semantic fiber to the same $E_{\mathrm{sem}}$ used in Stage~A and learns only $W'$. The sequence-independent matrix
\begin{equation}
    M[\phi(n,q),\, v]
    = E_{\mathrm{sem}}[v, n]\, W'[q, v],
    \qquad
    M\in\R^{d_x d_t\times|V|},
\end{equation}
is formed first. The logits are then one product,
\begin{equation}
    L_{\mathrm{flat}} = X^{\mathrm{norm}}_{\mathrm{flat}} M,
    \qquad
    X^{\mathrm{norm}}_{\mathrm{flat}}
    = \mathrm{reshape}\big(X^{\mathrm{norm}},\, BNT \times d_x d_t\big),
\end{equation}
viewed back to $(B, N, T, |V|)$. Because $E_{\mathrm{sem}}[\texttt{padding\_idx}]=0$, the padding column of $M$ is identically zero, and the padding logit is zero. A padding argmax on the generation path is reported as a warning. The logits are left unchanged, and no permanent mask is written onto the vocabulary axis.

Softmax over $v$ is a diagnostic map used for sampling or inspection. The training loss does not materialize it and does not materialize a one-hot target. Given the full window $\texttt{token\_ids\_full}$ of shape $(B, T{+}1, N)$,
\begin{align}
    \texttt{token\_ids} &= \texttt{token\_ids\_full}[:,\, 0{:}T,\, :], \\
    Y &= \texttt{token\_ids\_full}[:,\, 1{:}T{+}1,\, :],
\end{align}
so $Y[b,t,x]$ is the token of $S(t{+}1)$ and $Y$ has shape $(B, T, N)$. The loss is the mean token cross-entropy of the integer targets after the axis bridge,
\begin{equation}
    \mathcal{L}
    = \mathrm{CrossEntropy}\big(
        \mathrm{reshape}(L,\, BNT\times|V|),\;
        \mathrm{reshape}(Y_{\mathrm{perm}},\, BNT)
      \big),
\end{equation}
with ignore index equal to $\texttt{padding\_idx}$. The log-softmax sits inside that reduction. Ignore-index removal drops padding targets from $\mathcal{L}$. It does not replace the key mask of \cref{sec:steps25}, and the zero padding row of $E_{\mathrm{sem}}$ does not replace it either: that row only stops a padding identifier from contributing an embedding gradient.

The recorded last-slice score uses the final sentence axis of the same logits, $L[:, :, T{-}1, :]$, against $Y$'s last sentence, again ignoring padding. For the reported windows, $T=4$, so this slice is $S(4)$.

Gradient descent accumulates into $E_{\mathrm{sem}}$ along both the Stage~A path and the tied readout. $W'$ receives a gradient only from the readout.

\subsection{End-to-end training procedure}

One training example of shape $(B, T{+}1, N)$ is processed as follows.
\begin{enumerate}
    \item Reject the example unless its shape is $(B, T{+}1, N)$ with $B\ge 1$.
    \item Split it into the input prefix $\texttt{token\_ids}$ of shape $(B, T, N)$ and the shifted target $Y$ of shape $(B, T, N)$.
    \item Build $\texttt{pad\_mask}$ from $\texttt{token\_ids}$ by the permutation in \cref{sec:steps25}.
    \item Build $X$ by \cref{eq:step1}.
    \item For each of the $L$ layers, replace $X$ by $X^{\mathrm{out}}$ from \cref{sec:preln}, reusing $\texttt{pad\_mask}$.
    \item Apply $\mathrm{final\_norm}$ on $\phi(n,q)$.
    \item Form $M$ from $E_{\mathrm{sem}}$ and $W'$, then the logits $X^{\mathrm{norm}}_{\mathrm{flat}} M$.
    \item Permute $Y$ from $(B, T, N)$ to $(B, N, T)$ and evaluate the ignore-index cross-entropy against the logits.
\end{enumerate}
Teacher forcing uses ground-truth sentences on the input side for every term of $\mathcal{L}$. A predicted sentence is substituted for a ground-truth sentence only in the diagnostic multi-step rollout, which does not propagate a gradient. At evaluation time the same forward procedure is used with dropout disabled.

\begin{table}[t]
\centering
\small
\caption{Fiber role through one reduced-to-practice forward pass. The pair $(n,q)$ is eliminated at the readout and does not appear in $\mathcal{L}$.}
\label{tab:fibers}
\begin{tabular}{@{}lll@{}}
\toprule
Stage & Tensor & Storage shape \\
\midrule
Step 1 & $X$ & $(B, N, T, d_x, d_t)$ \\
Steps 2--5, summed over heads & $\mathrm{Attn}$ & $(B, N, T, d_x, d_t)$ \\
After both residual adds & $X^{\mathrm{out}}$ & $(B, N, T, d_x, d_t)$ \\
Readout-boundary norm & $X^{\mathrm{norm}}$ & $(B, N, T, d_x, d_t)$ \\
Tied readout & logits & $(B, N, T, |V|)$ \\
Shifted targets & $Y$ & $(B, T, N)$ \\
\bottomrule
\end{tabular}
\end{table}

\subsection{Faithfulness checks}
\label{sec:faithfulness}

The following conditions are part of the embodiment. Each one is a case in which a tensor can have a plausible shape and still implement a different map.
\begin{enumerate}
    \item The activation layout is $(B, N, T, d_x, d_t)$. The layout $(B, N, d_x, T, d_t)$ coincides with it whenever $T=d_x$.
    \item $\phi(n,q)=n\, d_t+q$ is shared by the weight view, the MLP, $\mathrm{final\_norm}$, and $M$. A swapped fiber $(d_t, d_x)$ has the same product.
    \item $P$ broadcasts over $t$ and $q$. $T^{\mathrm{emb}}$ broadcasts over $n$. The two broadcasts are not interchangeable.
    \item $\texttt{pad\_mask}$ is computed from token identifiers. $X_{\mathrm{pad}}=P+T^{\mathrm{emb}}$ is nonzero, so a mask inferred from $\|X\|$ selects the wrong slots.
    \item Key scores of padding slots are filled with the smallest finite value of the score dtype. Query slots are unmasked. The MLP receives no mask.
    \item $\texttt{pad\_mask}$ is local to one forward call and is shared by the layers of that call.
    \item Each head uses a full $(d_x, d_t)$ fiber. Heads are summed, with no factor $1/\sqrt{H}$.
    \item LayerNorm is applied to each branch input and, once more, to the final residual. The residual addends are the unnormalized stream.
    \item Dropout with a single rate $p$ is applied to the attention branch and to the MLP branch only.
    \item $M$ is materialized from $E_{\mathrm{sem}}$ and $W'$ before multiplication by $X$. The alternative intermediate of shape $BNT\times d_t\times|V|$ is the same function at much larger activation cost and is not the implemented order.
    \item The loss permutes $Y$ from $(B, T, N)$ to $(B, N, T)$ and drops padding targets by the ignore index. The key mask remains in force on the forward pass.
    \item Training compares logits to integer targets. The vocabulary softmax and the one-hot target are available as diagnostics and are not called by the training step.
\end{enumerate}

\section{Reference Implementation}
\label{sec:reference-code}

The listings below are the design operations of \texttt{four\_tensor\_model.py}, in pipeline order. Shape and range checks, the optimizer wrapper, and diagnostic warnings are omitted. Every arithmetic line is the source statement. Comments are shortened to ASCII.

\subsection{Input construction}

\begin{lstlisting}[style=pydesign]
def _sinusoidal_pe(positions, dim):
    # even 2i -> sin(pos / 10000^{2i/dim}); odd 2i+1 -> cos; leftover slot stays 0
    half = dim // 2
    i = torch.arange(half, device=positions.device, dtype=positions.dtype)
    div = 10000.0 ** (2 * i / dim)
    angle = positions.unsqueeze(-1) / div
    pe = positions.new_zeros(*positions.shape, dim)
    pe[..., 0:2 * half:2] = torch.sin(angle)
    pe[..., 1:2 * half:2] = torch.cos(angle)
    return pe

# Stage A dictionaries. padding_idx row is zero and receives no gradient.
self.E_sem = nn.Embedding(vocab_size, d_x, padding_idx=padding_idx)
self.E_temp = nn.Embedding(vocab_size, d_t, padding_idx=padding_idx)

# Stage B: P_xn, phase = x. Stage D: Temb, phase = x + t.
x_idx = torch.arange(N, dtype=torch.float32)
self.register_buffer("P", _sinusoidal_pe(x_idx, d_x), persistent=False)
tau_idx = torch.arange(T, dtype=torch.float32)
phase = x_idx[:, None] + tau_idx[None, :]          # (N, T)
self.register_buffer("Temb", _sinusoidal_pe(phase, d_t), persistent=False)

def build_token_tensor(self, token_ids):
    # token_ids (B, T, N) -> S (B, N, T, d_x, d_t)
    e_n = self.E_sem(token_ids)                     # (B, T, N, d_x)
    e_q = self.E_temp(token_ids)                    # (B, T, N, d_t)
    S = e_n.unsqueeze(-1) * e_q.unsqueeze(-2)       # outer product (n, q)
    return S.permute(0, 2, 1, 3, 4).contiguous()    # (B, N, T, d_x, d_t)

def build_input_tensor(self, token_ids):
    S = self.build_token_tensor(token_ids)
    S_tilde = S + self.P[None, :, None, :, None]    # broadcast over (t, q)
    return S_tilde + self.Temb[None, :, :, None, :] # broadcast over n
\end{lstlisting}

\subsection{Rank-4 attention head}

\begin{lstlisting}[style=pydesign]
def _rank4_weight(d_x_out, d_x_in, d_t_out, d_t_in):
    # Xavier on the flat (m p, n q) matrix, then view to (m, p, n, q).
    mat = torch.empty(d_x_out * d_t_out, d_x_in * d_t_in)
    nn.init.xavier_uniform_(mat)
    return nn.Parameter(mat.view(d_x_out, d_t_out, d_x_in, d_t_in).contiguous())

self.scale = 1.0 / math.sqrt(d_x_out * d_t_out)
self.W_Q = _rank4_weight(d_x_out, d_x, d_t_out, d_t)
self.W_K = _rank4_weight(d_x_out, d_x, d_t_out, d_t)
self.W_V = _rank4_weight(d_x_out, d_x, d_t_out, d_t)

def _contract(self, X, W):
    # Y[b,x,t,m,p] = sum_{n,q} W[m,p,n,q] * X[b,x,t,n,q]
    B, N, T, d_x_in, d_t_in = X.shape
    d_x_out, d_t_out, _, _ = W.shape
    X_flat = X.reshape(B * N * T, d_x_in * d_t_in)
    W_flat = W.view(d_x_out * d_t_out, d_x_in * d_t_in)
    return F.linear(X_flat, W_flat).view(B, N, T, d_x_out, d_t_out)

def compute_scores(self, Q, K):
    # Scores (B, N, T, N, T), joint over the key grid (x', t').
    B, N, T, d_x_out, d_t_out = Q.shape
    Qf = Q.reshape(B, N * T, d_x_out * d_t_out)
    Kf = K.reshape(B, N * T, d_x_out * d_t_out)
    scores = self.scale * torch.bmm(Qf, Kf.transpose(1, 2))
    return scores.view(B, N, T, N, T)

def joint_softmax(Scores, pad_mask=None):
    # Key axes only. finfo.min, so an all-PAD key row stays finite and uniform.
    B, N, T, _, _ = Scores.shape
    if pad_mask is not None:
        neg = torch.finfo(Scores.dtype).min
        Scores = Scores.masked_fill(pad_mask[:, None, None, :, :], neg)
    return F.softmax(Scores.reshape(B, N, T, N * T), dim=-1).view(B, N, T, N, T)

def attention_output(self, A, V):
    B, N, T, d_x_out, d_t_out = V.shape
    Attn_flat = torch.bmm(A.reshape(B, N * T, N * T),
                          V.reshape(B, N * T, d_x_out * d_t_out))
    return Attn_flat.view(B, N, T, d_x_out, d_t_out)

def forward(self, X, pad_mask):
    Q, K, V = self.build_qkv(X)          # three _contract calls
    A = self.joint_softmax(self.compute_scores(Q, K), pad_mask)
    return self.attention_output(A, V)
\end{lstlisting}

\subsection{Multi-head sum, Pre-LN, and the MLP}

\begin{lstlisting}[style=pydesign]
def forward(self, X, pad_mask):
    # Each head keeps a full (d_x, d_t) fiber. No 1/sqrt(H) factor.
    Attn = self.heads[0](X, pad_mask)
    for head in self.heads[1:]:
        Attn = Attn + head(X, pad_mask)
    return Attn

def compute_dff(d_x, d_t):
    return 4 * d_x * d_t

# d_in = d_x * d_t;  W1 (d_ff, d_in), W2 (d_in, d_ff)
self.fc1 = nn.Linear(d_in, d_ff)
self.fc2 = nn.Linear(d_ff, d_in)
self.ln = nn.LayerNorm(d_in)            # MLP branch, distinct from ln_attn
self.dropout = nn.Dropout(dropout)

def residual_after_attention(self, X, Attn):
    return X + self.dropout(Attn)

def mlp_block(self, X_out_attn):
    B, N, T, d_x, d_t = X_out_attn.shape
    v = self.ln(X_out_attn.reshape(B, N, T, d_x * d_t)).reshape(B * N * T, d_x * d_t)
    out = self.fc2(F.gelu(self.fc1(v)))
    return out.view(B, N, T, d_x, d_t)

def residual_after_mlp(self, X_out_attn, Z):
    return X_out_attn + self.dropout(Z)

def _norm_attn(self, X):
    B, N, T, d_x, d_t = X.shape
    return self.ln_attn(X.reshape(B, N, T, d_x * d_t)).reshape(B, N, T, d_x, d_t)

def forward(self, X, pad_mask):
    # Pre-LN on the attention branch; the MLP pre-norm is inside mlp_block.
    Attn = self.attention(self._norm_attn(X), pad_mask)
    return self.mlp(X, Attn)
\end{lstlisting}

\subsection{Tied readout, shifted loss, and the model step}

\begin{lstlisting}[style=pydesign]
self.W_prime = nn.Parameter(torch.empty(d_t, vocab_size))
nn.init.xavier_uniform_(self.W_prime)
self.final_norm = nn.LayerNorm(d_x * d_t)

def readout_logits_tied(self, X_output, E_sem):
    # M[n * d_t + q, v] = E_sem[v, n] * W_prime[q, v], then L = X_flat @ M.
    B, N, T, d_x, d_t = X_output.shape
    d = d_x * d_t
    X_flat = X_output.reshape(B * N * T, d)
    M = torch.einsum("vn,qv->nqv", E_sem, self.W_prime).reshape(d, self.vocab_size)
    return (X_flat @ M).view(B, N, T, self.vocab_size)

def build_pad_mask(self, token_ids):
    # token_ids (B, T, N) -> mask (B, N, T), True where the id is padding.
    return (token_ids == self.padding_idx).permute(0, 2, 1)

def forward(self, token_ids):
    pad_mask = self.build_pad_mask(token_ids)
    X = self.input_construction(token_ids)
    for layer in self.layers:
        X = layer(X, pad_mask)
    B, N, T, d_x, d_t = X.shape
    X_norm = self.final_norm(X.reshape(B, N, T, d_x * d_t)).reshape(B, N, T, d_x, d_t)
    return self.readout(X_norm, self.input_construction.E_sem.weight)

def build_targets(token_ids_full):
    # (B, T+1, N) -> Y (B, T, N) = S(1), ..., S(T)
    return token_ids_full[:, 1:, :].contiguous()

def cross_entropy_loss(logits, Y, pad_id):
    # logits (B, N, T, |V|), Y (B, T, N). The permute is the axis bridge.
    tgt = Y.permute(0, 2, 1).reshape(-1)
    return F.cross_entropy(logits.reshape(-1, logits.size(-1)), tgt,
                           ignore_index=pad_id)

def last_slice_token_stats(logits, token_ids_full, pad_id):
    # Reported score: non-padding tokens of S(T) only.
    last = logits[:, :, -1, :]
    gold = token_ids_full[:, -1, :]
    mask = gold != pad_id
    ce_sum = F.cross_entropy(last.reshape(-1, last.size(-1)), gold.reshape(-1),
                             ignore_index=pad_id, reduction="sum")
    n_correct = ((last.argmax(dim=-1) == gold) & mask).sum()
    n_tok = mask.sum()
    return ce_sum, n_correct, n_tok

def training_step(self, token_ids_full):
    token_ids = token_ids_full[:, :-1, :]
    logits = self.forward(token_ids)
    Y = build_targets(token_ids_full)
    loss = cross_entropy_loss(logits, Y, pad_id=self.padding_idx)
    return loss, logits
\end{lstlisting}

\section{4-Tensor Attention Backward Pass and Weight Update Rules}
\label{sec:backward_pass}

The derivation of the backward pass through the rank-4 attention mechanism requires rigorous tracking of the 4-dimensional index contractions. Because the token-position axes ($x, x', y, z$) and the temporal-slice axes ($t, t', u, w$) interact across multiple nested summations, we compute the exact analytical gradients traversing backward from the raw attention output $y_{xm;tp}$ through the joint softmax to the primary weight tensors. The temporal-context fiber is the index $q$.

\subsection{Joint Softmax Jacobian and Score Tensor Gradient}

Let $A_{yu,\, y'u'}$ denote the joint softmax probabilities computed over the score tensor $E_{yu,\,\tilde{y}\tilde{u}}$. The Jacobian of this transformation with respect to the pre-softmax scores evaluates to:
\begin{equation}
    \frac{\partial A_{yu,\,y'u'}}{\partial E_{xt,\,x't'}} = \delta_{xy}\delta_{tu}\, A_{xt,\,y'u'}\left(\delta_{x'y'}\delta_{t'u'} - A_{xt,\,x't'}\right)
\end{equation}
The Kronecker deltas $\delta_{xy}\delta_{tu}$ dictate that $A$ depends on $E$ exclusively through the row $E_{xt,\,\cdot}$ sharing the identical query index $(x,t) = (y,u)$. The joint-softmax rows remain entirely algebraically independent.

Applying the chain rule, we substitute this Jacobian to find the gradient of the loss with respect to the score tensor, denoted $\nabla E_{xt,\,x't'} \equiv \frac{\partial \mathcal{L}}{\partial E_{xt,\,x't'}}$. The deltas collapse the outer summation over $(y,u)$, yielding the exact row-wise softmax-gradient identity:
\begin{equation}
    \nabla E_{xt,\,x't'} = A_{xt,\,x't'} \left( \frac{\partial \mathcal{L}}{\partial A_{xt,\,x't'}} - \sum_{y',u'} \frac{\partial \mathcal{L}}{\partial A_{xt,\,y'u'}}\, A_{xt,\,y'u'} \right)
\end{equation}
For each query $(x,t)$, the gradient subtracts the $A$-weighted average of the incoming gradient across the entire key/value grid, subsequently scaled by the local probability $A_{xt,\,x't'}$.

\subsection{Gradients with Respect to $Q$, $K$, and $V$}

Recall the forward definition of the score tensor: $E_{xt,\,x't'} = \frac{1}{\sqrt{d_x d_t}} \sum_{m,p} Q_{xm;tp}\, K_{x'm;t'p}$. Differentiating $\mathcal{L}$ with respect to the query tensor $Q$ yields:
\begin{equation}
    \nabla Q_{zm;wp} \equiv \frac{\partial \mathcal{L}}{\partial Q_{zm;wp}} = \sum_{xt,\,x't'} \nabla E_{xt,\,x't'} \left( \frac{1}{\sqrt{d_x d_t}}\, \delta_{xz}\delta_{tw}\, K_{x'm;t'p} \right)
\end{equation}
Collapsing the deltas, we express the gradients for $Q$ and $K$ in slice-matrix notation, where each spatial-temporal position acts as a matrix:
\begin{align}
    \nabla Q_{z;w} &= \frac{1}{\sqrt{d_x d_t}} \sum_{x',t'} \nabla E_{zw,\,x't'}\, K_{x';t'} \\
    \nabla K_{z;w} &= \frac{1}{\sqrt{d_x d_t}} \sum_{x,t} \nabla E_{xt,\,zw}\, Q_{x;t}
\end{align}
This explicitly reveals the index asymmetry: $\nabla Q$ sums $\nabla E$ over its \emph{second} index pair with the first pair fixed, whereas $\nabla K$ sums over its \emph{first} index pair with the second fixed. This structural asymmetry directly mirrors their respective roles as row and column determinants of the score matrix.

Because the value tensor $V$ bypasses the score computation and enters exclusively at the final attention contraction ($y_{xm;tp} = \sum_{x',t'} A_{xt,\,x't'} V_{x'm;t'p}$), its gradient relies directly on the incoming upstream gradient $\nabla y_{xm;tp}$:
\begin{equation}
    \nabla V_{z;w} = \sum_{x,t} A_{xt,\,zw}\, \nabla y_{x;t}
\end{equation}
This operation is the exact transpose of the forward attention contraction: the backward pass distributes $\nabla y$ into $\nabla V$ using $A$ as weights over the query sequence $(x,t)$ for each key/value position $(z,w)$.

\subsection{Chain Rule to the Projection Weights}

The upstream gradient $\nabla y_{xm;tp}$ arrives at the attention block via the network's backward pass through the vocabulary readout, LayerNorm, MLP block, and sequential residual connections. Treating $\nabla Q, \nabla K, \nabla V$ as the explicitly derived downstream targets, we backpropagate into the learned weight tensors $W^Q, W^K, W^V \in \mathbb{R}^{d_x \times d_x \times d_t \times d_t}$.

$Q$, $K$, and $V$ are formed from $U = \mathrm{LN}(X)$. The forward projection (e.g., $Q_{xm;tp} = \sum_{n,q} W^Q_{mn,pq} U_{xn;tq}$) is a strict rank-4 bilinear contraction. Differentiating this contraction strips out $U$, yielding the analytical gradients for the network weights:
\begin{align}
    \frac{\partial \mathcal{L}}{\partial W^Q_{mn,pq}} &= \sum_{x,t} \nabla Q_{xm;tp}\, U_{xn;tq} \\
    \frac{\partial \mathcal{L}}{\partial W^K_{mn,pq}} &= \sum_{x,t} \nabla K_{xm;tp}\, U_{xn;tq} \\
    \frac{\partial \mathcal{L}}{\partial W^V_{mn,pq}} &= \sum_{x,t} \nabla V_{xm;tp}\, U_{xn;tq}
\end{align}
Each formulation is a tensor outer product between the upstream gradient (in the $m, p$ space) and $U$ (in the $n, q$ space), summed over positions $(x, t)$. The gradient continues from $U$ to the residual $X$ through the LayerNorm. 

\subsection{Gradient-Descent Update Rule}

The display is plain gradient descent, for exposition; training uses AdamW (\cref{sec:protocol}). Given a learning rate $\eta$ and the cross-entropy $\mathcal{L}$, the joint parameter updates are:
\begin{align}
    W^Q_{mn,pq} &\leftarrow W^Q_{mn,pq} - \eta\, \frac{\partial \mathcal{L}}{\partial W^Q_{mn,pq}} \\
    W^K_{mn,pq} &\leftarrow W^K_{mn,pq} - \eta\, \frac{\partial \mathcal{L}}{\partial W^K_{mn,pq}} \\
    W^V_{mn,pq} &\leftarrow W^V_{mn,pq} - \eta\, \frac{\partial \mathcal{L}}{\partial W^V_{mn,pq}}
\end{align}

\section{Comparative Baseline: Standard 1-Dimensional Autoregressive Attention}
\label{sec:baseline}

To rigorously evaluate the efficacy of the rank-4 structural decomposition, we establish a standard autoregressive (AR) transformer baseline tasked with the identical generative objective: predicting the future semantic state $\hat{S}(t)$ exclusively from the historical context window. 

For an observation window of $T=3$, the baseline model receives the exact same sequence of contextual states $\{S(0), S(1), S(2)\}$ and must generate the subsequent state $S(3)$. Crucially, to maintain a fair comparison and prevent causal leakage, the ground truth target $S(3)$ is strictly isolated from the input, mirroring the exact forward-prediction constraints imposed on the 4-tensor framework.

While the 4-tensor architecture explicitly decomposes the causal surface into independent token-position ($x$), temporal ($t$), semantic ($n$), and contextual ($q$) axes, a standard AR transformer fundamentally lacks this geometric capacity. Instead, it must concatenate the discrete semantic states into a single, flattened 1-D token sequence. The standard self-attention mechanism ($Q, K, V$) operates globally over this unidimensional string, relying entirely on 1-D positional encodings to implicitly approximate temporal boundaries. 

Consequently, this baseline serves not as an ablation of the 4-tensor mechanism, but as a definitive test of representation: it isolates the structural limits of applying standard flat-sequence attention to a problem domain that inherently demands spatial-temporal and semantic-contextual decomposition.

\subsection{1-D Sequence Packing and Standard QKV Architecture}

To process the structured context $X(2)$ through a standard transformer, the distinct semantic states must be collapsed into a contiguous 1-D token sequence. We eliminate internal padding tokens and concatenate the discrete sentences in temporal reading order. Because standard vocabularies natively encode punctuation, explicit separator tokens between states are unnecessary.

To illustrate this structural flattening, consider a simplified contextual sequence:
\begin{itemize}
    \item[] \textbf{Context $X(2)$:} \texttt{[BOS] tom went to the store . he bought milk . then he walked home .  [EOS]}
    \item[] \textbf{Target $S(3)$:} \texttt{[BOS] tom drank the milk . [EOS]}
\end{itemize}

The baseline architecture operates entirely over this flattened 1-D string. The input representations are constructed by summing the shared lexical embeddings with standard 1-D absolute positional encodings:
\begin{equation}
    H_i = E[v_i] + \mathrm{PE}(i)
\end{equation}
where $v_i$ represents the token ID at sequence position $i$. The temporal progression is processed via the standard causal self-attention mechanism:
\begin{equation}
    Y = \mathrm{softmax}\left(\frac{(HW^Q)(HW^K)^\top}{\sqrt{d}} + M_{\text{causal}}\right) (HW^V)
\end{equation}

The baseline follows the same structural policy as the 4-tensor model: projection weights $W^{Q,K,V} \in \mathbb{R}^{d \times d}$, MLP width $4d$, and dropout on both residual branches. LayerNorm is applied to the residual stream before attention and before the MLP, and again on the flattened fiber immediately before the tied vocabulary projection $L = H^{\text{norm}} E^\top$.

During inference, the model is seeded solely with the flattened context sequence \texttt{[BOS] + $X(2)$}. It must then autoregressively emit the predicted sequence $\hat{S}(3)$ step-by-step, appending its own generated tokens until an \texttt{[EOS]} token is produced. Consequently, the baseline processes $X(2)$ as a structurally undifferentiated context, relying entirely on the 1-D causal mask and positional encodings to implicitly approximate the explicit spatial-temporal and semantic-contextual boundaries geometrically enforced by the 4-tensor model.

\subsection{Informational Parity via Free-Running Autoregressive Decoding}

To ensure a rigorous evaluation, the baseline must operate under the exact same informational premise as the 4-tensor framework: the model is conditioned exclusively on the historical context $X(2)$, with zero exposure to the ground truth tokens of the target state $S(3)$ during inference.

In the 4-tensor architecture, the future state is decoded in a single, parallel forward pass across all spatial positions $x$:
\begin{equation}
    \hat{S}(3)_x = f\big(X(2)\big) \quad \text{for all } x \in \{0, \dots, N-1\}
\end{equation}

By contrast, the standard 1-D baseline approximates the target state sequentially. To maintain strict informational parity, this generation must be executed in a \emph{free-running} autoregressive mode. Let $C$ denote the flattened context sequence comprising $X(2)$ and the explicit \texttt{[BOS]} token that initiates the target state. The baseline generates the target sequence $\hat{S}(3) = [\hat{s}_1, \hat{s}_2, \dots]$ step-by-step by continuously appending its own previous predictions:
\begin{align}
    \text{Step 1: } & C \;\longrightarrow\; \hat{s}_1 \\
    \text{Step 2: } & C \oplus [\hat{s}_1] \;\longrightarrow\; \hat{s}_2 \\
    \text{Step 3: } & C \oplus [\hat{s}_1, \hat{s}_2] \;\longrightarrow\; \hat{s}_3 \\
    &\;\;\vdots \nonumber \\
    \text{Step } k\text{: } & C \oplus [\hat{s}_1, \dots, \hat{s}_{k-1}] \;\longrightarrow\; \hat{s}_k
\end{align}
where $\oplus$ denotes 1-D sequence concatenation. 

Because the prior predictions $\hat{s}_{<k}$ are deterministically generated from $C$ and the model parameters, they contain no external information regarding the ground truth target. If the model makes an erroneous prediction at step 2, it is strictly forced to condition on that error at step 3. This constraint precisely isolates the fundamental architectural distinction---single-shot parallel structural projection versus 1-D sequential compounding factorization---without confounding the evaluation through asymmetric access to the target sequence.

\subsection{Architectural Alignment and Capacity Matching}

The baseline width is mapped from the 4-tensor fibers. Its model dimension is the joint product $d = d_x d_t$, and the feed-forward width is $d_{ff} = 4d$. The two models use $H$ differently. The 4-tensor model adds full-size heads; the baseline splits $d$ across its heads. The runs in \cref{sec:experiment-data} use $H \in \{2,4\}$. Among them, $H=4$ and $L=2$ is the closest match in total parameter count.

\begin{table}[t]
\centering
\small
\caption{How a head count enters the parameter count. $d = d_x d_t$. Biases and LayerNorm are omitted. The baseline attention count is the reported case $H>1$, three maps of size $d\times d$ plus $W^O$. The 4-tensor count is three full-fiber maps on each of $H$ heads and no output projection. The 4-tensor token term is $E_{\mathrm{sem}}$, $E_{\mathrm{temp}}$, and the readout factor $W'$.}
\label{tab:heads}
\begin{tabular}{@{}lcc@{}}
\toprule
& 4-tensor & Baseline \\
\midrule
Width of each head & full $(d_x, d_t)$ fiber, $d_x d_t$ & $d/H$ \\
Head merge & sum, no output projection & concatenation, then $W^O$ \\
Attention parameters per layer & $3Hd^2$ & $4d^2$ \\
MLP parameters per layer & $8d^2$ & $8d^2$ \\
Token table and readout & $|V|\,(d_x + 2d_t)$ & $(|V| + 2)\,d$, tied \\
\bottomrule
\end{tabular}
\end{table}

With $d = 2048$, raising $H$ from $2$ to $4$ adds $6d^2 \approx 25.2$M parameters per layer to the 4-tensor model and none to the baseline, which is why the baseline count is the same in the first two rows of \cref{tab:runs}. At $H = 4$, $L = 2$ the totals are close, but they are spent differently: the 4-tensor model puts $100.7$M into attention and $4.7$M into its token table, the baseline $33.6$M and $75.2$M.

The embedding tables differ in shape. The baseline instantiates one table $E \in \R^{|V| \times (d_x d_t)}$. The 4-tensor model factorizes that boundary into $E_{\text{sem}} \in \R^{|V| \times d_x}$ and $E_{\text{temp}}$.

Finally, because the standard baseline lacks the explicit spatial-grid bounding inherently provided by the 4-tensor's position index $x$, its vocabulary must be artificially expanded to include explicit \texttt{[BOS]} and \texttt{[EOS]} tokens. These boundary tokens are strictly required to control the start and termination of the 1-D sequence generation, a mechanism entirely obviated by the exact geometric decoding of the 4-tensor architecture.

\subsection{Baseline design pipeline}
\label{sec:baseline-pipeline}

The comparison model is a decoder-only causal transformer on one flattened token string. It receives the same stored window $(B, T{+}1, N)$ as the 4-tensor model and is asked for the same last sentence. It does not build a rank-4 tensor, a joint $(x,t)$ softmax, or a pair of fibers. Width, depth, dropout, and the optimizer are the copies stated in \cref{sec:protocol}: $d = d_x d_t$, $d_{ff} = 4d$, $L$ layers, residual dropout $p$ on both branches, AdamW at learning rate $3\times 10^{-4}$ and weight decay $0.01$.

\paragraph{Packing.}
Padding identifiers are deleted inside each sentence and the remaining tokens are concatenated in reading order. No separator is inserted; sentence boundaries remain only as whatever punctuation the tokenizer already emitted. Two identifiers are appended after the content vocabulary, $\texttt{bos}=|V|$ and $\texttt{eos}=|V|+1$, and the padding index is left unused by them. For a window whose context is $S(0),\ldots,S(T{-}1)$ and whose target sentence is $S(T)$,
\begin{align}
    \mathrm{prefix} &= [\texttt{bos}] \oplus \mathrm{strip}(S(0)) \oplus \cdots \oplus \mathrm{strip}(S(T{-}1)), \\
    \mathrm{gold} &= \mathrm{strip}(S(T)) \oplus [\texttt{eos}].
\end{align}
Each row is then right-padded with the padding index to the longest row in the batch, and the true lengths $\ell^{\mathrm{pre}}_b$ and $\ell^{\mathrm{gold}}_b$ are retained. The gold span is a target. It is not written into the prefix.

\paragraph{Token representation.}
One table $E\in\R^{(|V|+2)\times d}$ embeds every identifier, with the padding row held at zero. Position is a single integer $i$ along the packed string, encoded by the same sinusoid as \cref{sec:step1} with phase $i$ and width $d$:
\begin{equation}
    H^{(0)}_i = E[\mathrm{id}_i] + \mathrm{PE}(i).
\end{equation}
There is no column index $x$ and no sentence index inside the position code.

\paragraph{Causal block.}
A key at position $j$ is blocked when $j>i$, and every padding key is blocked. The blocked entries of the score are filled with the smallest finite value of the score dtype, the same guard as the 4-tensor key mask. Heads partition $d$. With $H=1$ the maps $W^Q,W^K,W^V\in\R^{d\times d}$ have no output projection and the score scale is $1/\sqrt{d}$. With $H>1$, $H$ must divide $d$, each head has width $d_h=d/H$, the scale is $1/\sqrt{d_h}$, and a learned map $W^O\in\R^{d\times d}$ recombines the concatenated heads. The reported runs use $H\in\{2,4\}$, so they include $W^O$. One layer is Pre-LN on both branches, with distinct LayerNorms of width $d$ and GELU in the MLP:
\begin{align}
    H &\leftarrow H + \mathrm{Dropout}_{p}\big(\mathrm{Attn}(\mathrm{LN}^{(1)}_{d}(H))\big), \\
    H &\leftarrow H + \mathrm{Dropout}_{p}\big(\mathrm{MLP}(\mathrm{LN}^{(2)}_{d}(H))\big).
\end{align}
The residual stream is the raw sum. After the last layer, one further $\mathrm{LN}_{d}$ produces $H^{\mathrm{norm}}$. Attention still runs on the full packed prefix; only the inputs of attention and of the MLP are normalized.

\paragraph{Readout at the last live position.}
The vocabulary projection is tied, $L_i = H^{\mathrm{norm}}_i E^{\top}$. In the free-running loop only the logit at the last content position of each row is consumed. Because the tied map has no cross-position term,
\begin{equation}
    \big(H^{\mathrm{norm}} E^{\top}\big)_{b,\, i_b}
    = H^{\mathrm{norm}}_{b,\, i_b}\, E^{\top},
    \qquad
    i_b = \ell_b - 1.
\end{equation}
The implementation therefore normalizes the full sequence, slices that one vector, and multiplies by $E^{\top}$. The earlier positions remain available as keys and values on the next step. The slice changes the activation size, not the function.

\paragraph{Free-running objective.}
Training and the reported comparison use the same loop. Start from the packed prefix. At step $k=0,\ldots,\ell^{\mathrm{gold}}_b-1$, read the logit of the last live position, add the cross-entropy against the gold token at that step, then append $\mathrm{argmax}$ of that logit. The appended identifier is detached, so the gradient stops at the logit that chose it and does not pass through the discrete token. If the argmax differs from the gold token, the next step conditions on that own token. Rows that have already exhausted their gold span are left inactive. Two reductions are accumulated. The selection loss averages every gold token, including \texttt{[EOS]}. The quantity placed beside the 4-tensor last-slice score averages only the content tokens of $S(T)$, dropping \texttt{[EOS]}, on the same non-padding denominator. Evaluation decodes greedily from the prefix alone, stops at \texttt{[EOS]} or after $N$ content tokens, and aligns that string to the gold content for the token accuracy in \cref{tab:runs}.

Teacher-forced training, which would place gold $S(T)$ on the input side of later steps, is a different premise. It is not the path implemented in the comparison model, and its cross-entropy is not the number in \cref{tab:runs}.

The 4-tensor model receives a loss on all $T$ target sentences of a window, of which only $S(T)$ is absent from the input; the baseline receives a loss only on the tokens of $S(T)$.

\subsection{Code for the fair comparison}
\label{sec:baseline-code}

Ordinary causal self-attention, Pre-LN, and the position-wise MLP are omitted. The statements below are the parts of \texttt{baseline\_ar\_attention\_model.py} that fix the comparison: the same window, a premise that excludes gold $S(T)$, a tied readout that is algebraically unchanged by the last-position slice, and a content-token denominator aligned with the 4-tensor last slice.

\begin{lstlisting}[style=pydesign]
def compute_d(d_x, d_t):
    return d_x * d_t                         # one width, d_ff = 4*d elsewhere

def bos_eos_ids(content_vocab_size):
    # Appended after the pool vocabulary. The padding index is not reused.
    return content_vocab_size, content_vocab_size + 1

def strip_pad(row, pad_id):
    return row[row != pad_id]

# token_ids_full (B, T+1, N). No separator token.
parts = [strip_pad(token_ids_full[b, t], padding_idx) for t in range(T + 1)]
prefix = torch.cat([bos, *parts[:T]])       # [BOS] + S(0) .. S(T-1)
gold = torch.cat([parts[T], eos])           # S(T) + [EOS], target only
\end{lstlisting}

\begin{lstlisting}[style=pydesign]
def forward_last(self, ids, lengths=None):
    # Full causal stack, then one vector into the tied table.
    # forward(ids)[b, i] == forward_last(ids, lengths)[b] at i = lengths[b]-1.
    H = self._encode(ids)
    if lengths is None:
        H_last = H[:, -1, :]
    else:
        idx = (lengths - 1).clamp(min=0, max=H.shape[1] - 1)
        H_last = H[torch.arange(H.shape[0], device=ids.device), idx]
    return F.linear(H_last, self.E.weight)
\end{lstlisting}

\begin{lstlisting}[style=pydesign]
# One free-running step. Gold is the target; the next input is the model's argmax.
last = self.forward_last(ids[:, :L_now], lengths)
tgt = packed.gold[:, step]
loss_sum = loss_sum + F.cross_entropy(last[active], tgt[active], reduction="sum")
is_content = active & (tgt != self.eos_id)   # drop EOS: same slots as 4-tensor S(T)
content_sum = content_sum + F.cross_entropy(
    last[is_content], tgt[is_content], reduction="sum")
hat = last.argmax(dim=-1).detach()
ids[active, write_at[active]] = hat[active]

def training_step(self, token_ids_full):
    return self.training_step_fair(token_ids_full)

# Eval string is (B, N): store content, skip EOS, stop after N tokens.
emit = live & (nxt != self.eos_id) & (n_content < self.N)
hat[rows, n_content[rows]] = nxt[rows]
g = packed.gold[b, : int(packed.gold_len[b])]
if g.numel() > 0 and int(g[-1]) == self.eos_id:
    g = g[:-1]                               # accuracy gold is S(T) without EOS
out[b, :n] = g[:n]
\end{lstlisting}

\section{Experimental Setup}
\label{sec:implementation}

The runs below train the Pre-LN 4-tensor model and the free-running 1-D baseline on the same semantic windows. Both place LayerNorm on the residual stream before attention and before the MLP, apply dropout on both residual branches, and apply one further LayerNorm on the fiber immediately before the tied readout. The number reported across the two models is the mean cross-entropy on the non-padding tokens of the last sentence.

\subsection{Corpus and windows}
\label{sec:corpus}

The corpus is the ROCStories pool of spring 2016, winter 2017, and the accompanying cloze files: 101{,}671 stories in total. Each stored window has shape $(T{+}1) \times N = 5 \times 16$. The model conditions on $ X(3) \leftarrow \{ S(0), S(1), S(2), S(3)\}.$ and the recorded score is the cross-entropy of $S(4)$. Sentences are padded to $N=16$ with padding index $0$; the pad fraction over the pool is $0.3876$, and $2{,}457$ sentences were truncated because they exceeded $N=16$ tokens. The vocabulary contains the $36{,}711$ tokens of frequency at least one. The vocabulary is built over the full pool, including the validation split. The baseline adds \texttt{[BOS]} and \texttt{[EOS]}, giving vocabulary size $36{,}713$.

Of these windows, $97{,}929$ are the training split and $3{,}742$ are the validation split. Early stopping and checkpoint selection use the validation split, and the tables below report that split. The validation denominator for $S(4)$ is $32{,}173$ non-padding tokens. A uniform distribution over the $36{,}711$-word vocabulary has cross-entropy $\ln 36711 \approx 10.51$.

\subsection{Optimization and checkpoint rule}
\label{sec:protocol}

Both models use $d_x = 64$ and $d_t = 32$. The baseline therefore has width $d = d_x d_t = 2048$ and feed-forward width $d_{ff} = 8192$. Dropout is $0.1$. The optimizer is AdamW with learning rate $3 \times 10^{-4}$ and weight decay $0.01$. Each GPU receives a batch of $32$; on two RTX 3080 GPUs the effective batch is $64$. Seeds $41$, $42$, and $43$ are fixed and deterministic algorithms are enabled. The head count and layer count of each setting are those of \cref{tab:runs}. The rollout runs of \cref{sec:protocol-rollout} use the same width, dropout, optimizer, and batch.

In both experiments, training stops when at least five epochs have passed since the minimum of that experiment's selection loss and the current value of that loss is above $1.01$ times the minimum, or when the epoch budget is exhausted. The reported checkpoint is the epoch of that minimum. The selection loss is defined separately for each experiment below. It is the quantity written to the validation-loss column of the log, and it is distinct from the sentence scores placed beside the models.

\begin{table}[t]
\centering
\caption{Shared setup of the runs. The selection row is the stopping arithmetic; the loss it minimizes is stated in \cref{sec:protocol-s4,sec:protocol-rollout}.}
\label{tab:setup}
\begin{tabular}{@{}lp{0.68\textwidth}@{}}
\toprule
Item & Value \\
\midrule
Corpus & ROCStories, five sentences stored as $5 \times 16$ \\
Split & $97{,}929$ train / $3{,}742$ validation \\
Vocabulary & $36{,}711$; baseline $36{,}713$ with \texttt{[BOS]}, \texttt{[EOS]} \\
Width & $d_x=64$, $d_t=32$; baseline $d=2048$, $d_{ff}=8192$ \\
Dropout & $0.1$ on both residual branches \\
Optimizer & AdamW, learning rate $3\times 10^{-4}$, weight decay $0.01$ \\
Batch & $32$ per GPU, effective batch $64$, two RTX 3080 \\
Selection & minimum validation loss; stop once five epochs have passed and the loss exceeds $1.01\times$ that minimum \\
\bottomrule
\end{tabular}
\end{table}

\subsubsection{Prediction of $S(4)$}
\label{sec:protocol-s4}

The model conditions on $X(3) \leftarrow \{ S(0), S(1), S(2), S(3) \}$. For the 4-tensor model the selection loss is the teacher-forced window cross-entropy on $S(1),\ldots,S(4)$, with padding ignored. For the baseline the selection loss is the free-running cross-entropy of its own $S(4)$ tokens, including \texttt{[EOS]}. The column placed beside both models is a different quantity: the token-weighted mean cross-entropy of the non-padding content tokens of $S(4)$. On the 4-tensor side this is one parallel forward pass, and the gold tokens of $S(4)$ are targets only. On the baseline side the same tokens are scored by free-running decoding from \texttt{[BOS]} plus the flattened context, with \texttt{[EOS]} removed from the denominator. The overlap terms on $S(1), S(2), S(3)$ add no gold tokens of $S(4)$ from outside the input, so both models receive the same story, $S(0)$ through $S(3)$, when they predict $S(4)$. Token accuracy is the argmax match rate on those same $32{,}173$ positions. \Cref{tab:runs} records the mean over seeds $41$, $42$, and $43$, and \cref{tab:runs-seeds} records each seed.

\subsubsection{Rollout prediction of $S(3)$ and $S(4)$}
\label{sec:protocol-rollout}

The same five-sentence records are read, and the window length is fixed at $T=3$. A checkpoint trained at $T=4$ is not reused: the temporal embedding is built for the training length, and the last slice of a $T=4$ model is time index $3$. $S(4)$ is not a training target in either model. Gold $S(3)$ and gold $S(4)$ are targets only; neither sentence is written into a context.

For the 4-tensor model the training input is $\{S(0), S(1), S(2)\}$ and the teacher-forced targets are $\{S(1), S(2), S(3)\}$. The selection loss is the mean cross-entropy of that window, with padding ignored. After the weights are held fixed, two forwards are scored separately. The first reads $\{S(0), S(1), S(2)\}$ and compares the last-slice argmax $\hat{S}(3)$ with $S(3)$. The second reads $\{S(1), S(2), \hat{S}(3)\}$ and compares the last-slice argmax $\hat{S}(4)$ with $S(4)$. Each score is the token-weighted mean cross-entropy of the non-padding tokens of that one sentence. The two means are logged apart and do not enter the selection loss.

For the baseline the training context is \texttt{[BOS]} followed by the non-padding tokens of $S(0)$, $S(1)$, and $S(2)$, under the causal mask. The training target is the non-padding tokens of $S(3)$ followed by \texttt{[EOS]}. Decoding is free-running: the tokens appended to the context are the model's own argmax, and the gold sentence is the loss target only. The selection loss is that free-running cross-entropy, including \texttt{[EOS]}. The first evaluation uses the same context and scores $\hat{S}(3)$ against $S(3)$. The second uses the non-padding tokens of $S(1)$, $S(2)$, and $\hat{S}(3)$ as context and scores $\hat{S}(4)$ against $S(4)$. Both reported cross-entropies drop \texttt{[EOS]} and divide by the non-padding content tokens of the sentence being scored. They are logged apart, and the checkpoint remains the minimum of the selection loss on $S(3)$ together with \texttt{[EOS]}.

\section{Experiment Data}
\label{sec:experiment-data}

\Cref{tab:runs} lists the matched Pre-LN settings at the checkpoint selected by the rule in \cref{sec:protocol-s4}. Each entry is written as 4-tensor / baseline. The first number is the mean over seeds $41$, $42$, and $43$, and the second is the sample standard deviation. \Cref{tab:runs-seeds} lists each seed. The difference on a seed is
\begin{equation}
    \Delta = \mathcal{L}^{\mathrm{AR}}_{S(4)} - \mathcal{L}^{4\mathrm{T}}_{S(4)} .
\end{equation}
A positive $\Delta$ means the 4-tensor cross-entropy is lower. The summary reports the mean and sample standard deviation of the three paired differences. Perplexity is $\exp(\mathcal{L}_{S(4)})$ and is omitted from the tables.

\begin{table}[t]
\centering
\footnotesize
\setlength{\tabcolsep}{3.5pt}
\caption{Matched settings at each seed's selected checkpoint. Each entry is 4-tensor / baseline: the mean and the sample standard deviation over seeds $41$, $42$, and $43$. $\mathcal{L}_{S(4)}$ is the mean cross-entropy on the $32{,}173$ non-padding tokens of the last sentence, and accuracy is the argmax match rate on the same tokens. $\Delta$ is the paired difference defined in the text.}
\label{tab:runs}
\vspace{8pt}
\begin{tabular}{@{}rrcccc@{}}
\toprule
$H$ & $L$ & parameters & $\mathcal{L}_{S(4)}$ & acc.\ (\%) & $\Delta$ \\
\midrule
2 & 2 & 122.2M / 175.9M & $5.580 \pm 0.017$ / $5.868 \pm 0.157$ & $16.4 \pm 0.6$ / $8.8 \pm 3.1$ & $0.288 \pm 0.165$ \\
4 & 2 & 172.5M / 175.9M & $5.545 \pm 0.010$ / $5.655 \pm 0.059$ & $16.1 \pm 0.9$ / $11.3 \pm 1.6$ & $0.111 \pm 0.050$ \\
4 & 3 & 256.4M / 226.2M & $5.536 \pm 0.005$ / $5.825 \pm 0.326$ & $16.0 \pm 0.5$ / $9.0 \pm 5.1$ & $0.290 \pm 0.321$ \\
\bottomrule
\end{tabular}
\end{table}

\begin{table}[t]
\centering
\footnotesize
\setlength{\tabcolsep}{4pt}
\caption{Each seed at its own selected checkpoint. Epoch is selected/last, written as 4-tensor / baseline. The other columns match \cref{tab:runs}.}
\label{tab:runs-seeds}
\vspace{6pt}
\begin{tabular}{@{}rrrrrrr@{}}
\toprule
$H$ & $L$ & seed & epoch & $\mathcal{L}_{S(4)}$ & acc.\ (\%) & $\Delta$ \\
\midrule
2 & 2 & 41 & 7/12 / 32/37 & 5.582 / 6.023 & 16.7 / 6.8 & 0.440 \\
2 & 2 & 42 & 4/9 / 54/61 & 5.596 / 5.708 & 15.8 / 12.4 & 0.112 \\
2 & 2 & 43 & 4/9 / 30/39 & 5.562 / 5.873 & 16.8 / 7.3 & 0.311 \\
4 & 2 & 41 & 4/9 / 23/28 & 5.544 / 5.676 & 17.0 / 13.1 & 0.131 \\
4 & 2 & 42 & 5/10 / 45/50 & 5.535 / 5.589 & 15.3 / 10.2 & 0.054 \\
4 & 2 & 43 & 6/11 / 45/50 & 5.554 / 5.701 & 16.0 / 10.5 & 0.147 \\
4 & 3 & 41 & 7/12 / 13/18 & 5.541 / 6.200 & 16.5 / 3.3 & 0.659 \\
4 & 3 & 42 & 6/11 / 29/35 & 5.531 / 5.667 & 15.4 / 11.1 & 0.136 \\
4 & 3 & 43 & 6/11 / 33/40 & 5.534 / 5.608 & 16.1 / 12.7 & 0.074 \\
\bottomrule
\end{tabular}
\end{table}

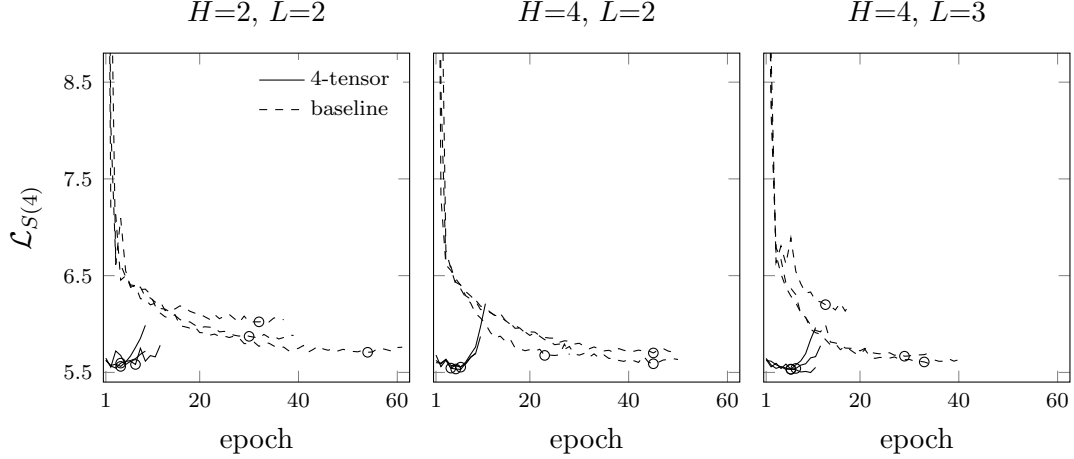
\begin{figure}[t]
\centering
\begin{tikzpicture}
\begin{groupplot}[
    group style={group size=3 by 1, horizontal sep=9pt},
    width=4.05cm,
    height=4.35cm,
    scale only axis,
    xmin=0.5, xmax=62.5,
    ymin=5.4, ymax=8.8,
    xlabel={epoch},
    xtick={1,20,40,60},
    ytick={5.5,6.5,7.5,8.5},
    tick label style={font=\scriptsize},
    label style={font=\small},
    title style={font=\small},
    legend style={font=\scriptsize, draw=none, at={(0.98,0.98)}, anchor=north east},
    legend cell align=left,
    clip=true,
]
\nextgroupplot[ylabel={$\mathcal{L}_{S(4)}$}, title={$H{=}2$, $L{=}2$}]
\addplot[black] table[x=epoch, y=s4_ce, col sep=comma] {figures/dynamics/h2l2_s41_4t.csv};
\addplot[forget plot, black] table[x=epoch, y=s4_ce, col sep=comma] {figures/dynamics/h2l2_s42_4t.csv};
\addplot[forget plot, black] table[x=epoch, y=s4_ce, col sep=comma] {figures/dynamics/h2l2_s43_4t.csv};
\addplot[black, dashed] table[x=epoch, y=s4_ce, col sep=comma] {figures/dynamics/h2l2_s41_ar.csv};
\addplot[forget plot, black, dashed] table[x=epoch, y=s4_ce, col sep=comma] {figures/dynamics/h2l2_s42_ar.csv};
\addplot[forget plot, black, dashed] table[x=epoch, y=s4_ce, col sep=comma] {figures/dynamics/h2l2_s43_ar.csv};
\addplot[forget plot, only marks, mark=o, mark size=1.8pt, black] table[x=epoch, y=s4_ce, col sep=comma] {figures/dynamics/h2l2_s41_4t_mark.csv};
\addplot[forget plot, only marks, mark=o, mark size=1.8pt, black] table[x=epoch, y=s4_ce, col sep=comma] {figures/dynamics/h2l2_s42_4t_mark.csv};
\addplot[forget plot, only marks, mark=o, mark size=1.8pt, black] table[x=epoch, y=s4_ce, col sep=comma] {figures/dynamics/h2l2_s43_4t_mark.csv};
\addplot[forget plot, only marks, mark=o, mark size=1.8pt, black] table[x=epoch, y=s4_ce, col sep=comma] {figures/dynamics/h2l2_s41_ar_mark.csv};
\addplot[forget plot, only marks, mark=o, mark size=1.8pt, black] table[x=epoch, y=s4_ce, col sep=comma] {figures/dynamics/h2l2_s42_ar_mark.csv};
\addplot[forget plot, only marks, mark=o, mark size=1.8pt, black] table[x=epoch, y=s4_ce, col sep=comma] {figures/dynamics/h2l2_s43_ar_mark.csv};
\legend{4-tensor, baseline}
\nextgroupplot[title={$H{=}4$, $L{=}2$}, yticklabels={}]
\addplot[black] table[x=epoch, y=s4_ce, col sep=comma] {figures/dynamics/h4l2_s41_4t.csv};
\addplot[forget plot, black] table[x=epoch, y=s4_ce, col sep=comma] {figures/dynamics/h4l2_s42_4t.csv};
\addplot[forget plot, black] table[x=epoch, y=s4_ce, col sep=comma] {figures/dynamics/h4l2_s43_4t.csv};
\addplot[black, dashed] table[x=epoch, y=s4_ce, col sep=comma] {figures/dynamics/h4l2_s41_ar.csv};
\addplot[forget plot, black, dashed] table[x=epoch, y=s4_ce, col sep=comma] {figures/dynamics/h4l2_s42_ar.csv};
\addplot[forget plot, black, dashed] table[x=epoch, y=s4_ce, col sep=comma] {figures/dynamics/h4l2_s43_ar.csv};
\addplot[forget plot, only marks, mark=o, mark size=1.8pt, black] table[x=epoch, y=s4_ce, col sep=comma] {figures/dynamics/h4l2_s41_4t_mark.csv};
\addplot[forget plot, only marks, mark=o, mark size=1.8pt, black] table[x=epoch, y=s4_ce, col sep=comma] {figures/dynamics/h4l2_s42_4t_mark.csv};
\addplot[forget plot, only marks, mark=o, mark size=1.8pt, black] table[x=epoch, y=s4_ce, col sep=comma] {figures/dynamics/h4l2_s43_4t_mark.csv};
\addplot[forget plot, only marks, mark=o, mark size=1.8pt, black] table[x=epoch, y=s4_ce, col sep=comma] {figures/dynamics/h4l2_s41_ar_mark.csv};
\addplot[forget plot, only marks, mark=o, mark size=1.8pt, black] table[x=epoch, y=s4_ce, col sep=comma] {figures/dynamics/h4l2_s42_ar_mark.csv};
\addplot[forget plot, only marks, mark=o, mark size=1.8pt, black] table[x=epoch, y=s4_ce, col sep=comma] {figures/dynamics/h4l2_s43_ar_mark.csv};
\nextgroupplot[title={$H{=}4$, $L{=}3$}, yticklabels={}]
\addplot[black] table[x=epoch, y=s4_ce, col sep=comma] {figures/dynamics/h4l3_s41_4t.csv};
\addplot[forget plot, black] table[x=epoch, y=s4_ce, col sep=comma] {figures/dynamics/h4l3_s42_4t.csv};
\addplot[forget plot, black] table[x=epoch, y=s4_ce, col sep=comma] {figures/dynamics/h4l3_s43_4t.csv};
\addplot[black, dashed] table[x=epoch, y=s4_ce, col sep=comma] {figures/dynamics/h4l3_s41_ar.csv};
\addplot[forget plot, black, dashed] table[x=epoch, y=s4_ce, col sep=comma] {figures/dynamics/h4l3_s42_ar.csv};
\addplot[forget plot, black, dashed] table[x=epoch, y=s4_ce, col sep=comma] {figures/dynamics/h4l3_s43_ar.csv};
\addplot[forget plot, only marks, mark=o, mark size=1.8pt, black] table[x=epoch, y=s4_ce, col sep=comma] {figures/dynamics/h4l3_s41_4t_mark.csv};
\addplot[forget plot, only marks, mark=o, mark size=1.8pt, black] table[x=epoch, y=s4_ce, col sep=comma] {figures/dynamics/h4l3_s42_4t_mark.csv};
\addplot[forget plot, only marks, mark=o, mark size=1.8pt, black] table[x=epoch, y=s4_ce, col sep=comma] {figures/dynamics/h4l3_s43_4t_mark.csv};
\addplot[forget plot, only marks, mark=o, mark size=1.8pt, black] table[x=epoch, y=s4_ce, col sep=comma] {figures/dynamics/h4l3_s41_ar_mark.csv};
\addplot[forget plot, only marks, mark=o, mark size=1.8pt, black] table[x=epoch, y=s4_ce, col sep=comma] {figures/dynamics/h4l3_s42_ar_mark.csv};
\addplot[forget plot, only marks, mark=o, mark size=1.8pt, black] table[x=epoch, y=s4_ce, col sep=comma] {figures/dynamics/h4l3_s43_ar_mark.csv};
\end{groupplot}
\end{tikzpicture}
\caption{Validation cross-entropy of the non-padding tokens of $S(4)$. \textbf{This is the sentence absent from the input:} the model reads $X(3)\leftarrow\{S(0),S(1),S(2),S(3)\}$, and $S(4)$ is not in that window. Each solid curve is one 4-tensor seed and each dashed curve is the paired free-running baseline, for seeds $41$, $42$, and $43$. Open circles are the selected checkpoints in \cref{tab:runs-seeds}. Epoch 1 of each baseline lies above the plotted range ($12.4$--$18.2$). Epoch 2 of the seed-$41$ baseline lies above it at $H{=}2$, $L{=}2$ ($10.05$) and at $H{=}4$, $L{=}2$ ($10.27$). The horizontal axis is the epoch index, not wall-clock time: at $H{=}4$, $L{=}2$ the 4-tensor runs took $2.0$--$2.4$ hours and the baseline runs took $25.1$--$45.2$ hours.}
\label{fig:dynamics}
\end{figure}

\begin{figure}[t]
\centering
\begin{tikzpicture}
\begin{groupplot}[
    group style={group size=3 by 2, horizontal sep=9pt, vertical sep=16pt},
    width=4.05cm,
    height=3.35cm,
    scale only axis,
    xmin=0.5, xmax=62.5,
    xlabel={},
    xtick={1,20,40,60},
    tick label style={font=\scriptsize},
    label style={font=\small},
    title style={font=\small},
    clip=true,
]
\nextgroupplot[ylabel={4-tensor}, title={$H{=}2$, $L{=}2$}, ymin=1.0, ymax=2.7, ytick={1.2,1.6,2.0,2.4}]
\addplot[black] table[x=epoch, y=train_ce, col sep=comma] {figures/dynamics/h2l2_s41_4t_train.csv};
\addplot[forget plot, black] table[x=epoch, y=train_ce, col sep=comma] {figures/dynamics/h2l2_s42_4t_train.csv};
\addplot[forget plot, black] table[x=epoch, y=train_ce, col sep=comma] {figures/dynamics/h2l2_s43_4t_train.csv};
\addplot[forget plot, only marks, mark=o, mark size=1.6pt, black] table[x=epoch, y=train_ce, col sep=comma] {figures/dynamics/h2l2_s41_4t_train_mark.csv};
\addplot[forget plot, only marks, mark=o, mark size=1.6pt, black] table[x=epoch, y=train_ce, col sep=comma] {figures/dynamics/h2l2_s42_4t_train_mark.csv};
\addplot[forget plot, only marks, mark=o, mark size=1.6pt, black] table[x=epoch, y=train_ce, col sep=comma] {figures/dynamics/h2l2_s43_4t_train_mark.csv};
\nextgroupplot[title={$H{=}4$, $L{=}2$}, yticklabels={}, ymin=1.0, ymax=2.7, ytick={1.2,1.6,2.0,2.4}]
\addplot[black] table[x=epoch, y=train_ce, col sep=comma] {figures/dynamics/h4l2_s41_4t_train.csv};
\addplot[forget plot, black] table[x=epoch, y=train_ce, col sep=comma] {figures/dynamics/h4l2_s42_4t_train.csv};
\addplot[forget plot, black] table[x=epoch, y=train_ce, col sep=comma] {figures/dynamics/h4l2_s43_4t_train.csv};
\addplot[forget plot, only marks, mark=o, mark size=1.6pt, black] table[x=epoch, y=train_ce, col sep=comma] {figures/dynamics/h4l2_s41_4t_train_mark.csv};
\addplot[forget plot, only marks, mark=o, mark size=1.6pt, black] table[x=epoch, y=train_ce, col sep=comma] {figures/dynamics/h4l2_s42_4t_train_mark.csv};
\addplot[forget plot, only marks, mark=o, mark size=1.6pt, black] table[x=epoch, y=train_ce, col sep=comma] {figures/dynamics/h4l2_s43_4t_train_mark.csv};
\nextgroupplot[title={$H{=}4$, $L{=}3$}, yticklabels={}, ymin=1.0, ymax=2.7, ytick={1.2,1.6,2.0,2.4}]
\addplot[black] table[x=epoch, y=train_ce, col sep=comma] {figures/dynamics/h4l3_s41_4t_train.csv};
\addplot[forget plot, black] table[x=epoch, y=train_ce, col sep=comma] {figures/dynamics/h4l3_s42_4t_train.csv};
\addplot[forget plot, black] table[x=epoch, y=train_ce, col sep=comma] {figures/dynamics/h4l3_s43_4t_train.csv};
\addplot[forget plot, only marks, mark=o, mark size=1.6pt, black] table[x=epoch, y=train_ce, col sep=comma] {figures/dynamics/h4l3_s41_4t_train_mark.csv};
\addplot[forget plot, only marks, mark=o, mark size=1.6pt, black] table[x=epoch, y=train_ce, col sep=comma] {figures/dynamics/h4l3_s42_4t_train_mark.csv};
\addplot[forget plot, only marks, mark=o, mark size=1.6pt, black] table[x=epoch, y=train_ce, col sep=comma] {figures/dynamics/h4l3_s43_4t_train_mark.csv};
\nextgroupplot[ylabel={baseline}, xlabel={epoch}, ymin=4.7, ymax=16.5, ytick={6,10,14}]
\addplot[black, dashed] table[x=epoch, y=train_ce, col sep=comma] {figures/dynamics/h2l2_s41_ar_train.csv};
\addplot[forget plot, black, dashed] table[x=epoch, y=train_ce, col sep=comma] {figures/dynamics/h2l2_s42_ar_train.csv};
\addplot[forget plot, black, dashed] table[x=epoch, y=train_ce, col sep=comma] {figures/dynamics/h2l2_s43_ar_train.csv};
\addplot[forget plot, only marks, mark=o, mark size=1.6pt, black] table[x=epoch, y=train_ce, col sep=comma] {figures/dynamics/h2l2_s41_ar_train_mark.csv};
\addplot[forget plot, only marks, mark=o, mark size=1.6pt, black] table[x=epoch, y=train_ce, col sep=comma] {figures/dynamics/h2l2_s42_ar_train_mark.csv};
\addplot[forget plot, only marks, mark=o, mark size=1.6pt, black] table[x=epoch, y=train_ce, col sep=comma] {figures/dynamics/h2l2_s43_ar_train_mark.csv};
\nextgroupplot[xlabel={epoch}, yticklabels={}, ymin=4.7, ymax=16.5, ytick={6,10,14}]
\addplot[black, dashed] table[x=epoch, y=train_ce, col sep=comma] {figures/dynamics/h4l2_s41_ar_train.csv};
\addplot[forget plot, black, dashed] table[x=epoch, y=train_ce, col sep=comma] {figures/dynamics/h4l2_s42_ar_train.csv};
\addplot[forget plot, black, dashed] table[x=epoch, y=train_ce, col sep=comma] {figures/dynamics/h4l2_s43_ar_train.csv};
\addplot[forget plot, only marks, mark=o, mark size=1.6pt, black] table[x=epoch, y=train_ce, col sep=comma] {figures/dynamics/h4l2_s41_ar_train_mark.csv};
\addplot[forget plot, only marks, mark=o, mark size=1.6pt, black] table[x=epoch, y=train_ce, col sep=comma] {figures/dynamics/h4l2_s42_ar_train_mark.csv};
\addplot[forget plot, only marks, mark=o, mark size=1.6pt, black] table[x=epoch, y=train_ce, col sep=comma] {figures/dynamics/h4l2_s43_ar_train_mark.csv};
\nextgroupplot[xlabel={epoch}, yticklabels={}, ymin=4.7, ymax=16.5, ytick={6,10,14}]
\addplot[black, dashed] table[x=epoch, y=train_ce, col sep=comma] {figures/dynamics/h4l3_s41_ar_train.csv};
\addplot[forget plot, black, dashed] table[x=epoch, y=train_ce, col sep=comma] {figures/dynamics/h4l3_s42_ar_train.csv};
\addplot[forget plot, black, dashed] table[x=epoch, y=train_ce, col sep=comma] {figures/dynamics/h4l3_s43_ar_train.csv};
\addplot[forget plot, only marks, mark=o, mark size=1.6pt, black] table[x=epoch, y=train_ce, col sep=comma] {figures/dynamics/h4l3_s41_ar_train_mark.csv};
\addplot[forget plot, only marks, mark=o, mark size=1.6pt, black] table[x=epoch, y=train_ce, col sep=comma] {figures/dynamics/h4l3_s42_ar_train_mark.csv};
\addplot[forget plot, only marks, mark=o, mark size=1.6pt, black] table[x=epoch, y=train_ce, col sep=comma] {figures/dynamics/h4l3_s43_ar_train_mark.csv};
\end{groupplot}
\end{tikzpicture}
\caption{Training cross-entropy on separate axes. The upper row is the 4-tensor model trained in these experiments. On a training window the input is $X(3) \leftarrow \{ S(0), S(1), S(2), S(3) \}$, and the loss is the mean cross-entropy of the shifted window $X(4) \leftarrow \{ S(1), S(2), S(3), S(4) \}$. The lower row is the baseline, whose loss is the free-running loss on $S(4)$, including [EOS]. Each curve is one of seeds $41$, $42$, and $43$. Open circles are the selected checkpoints in \cref{tab:runs-seeds}. Epoch 1 of each baseline lies above the plotted range ($29.1$--$32.6$). The horizontal axis is the epoch index, not wall-clock time.}
\label{fig:train}
\end{figure}
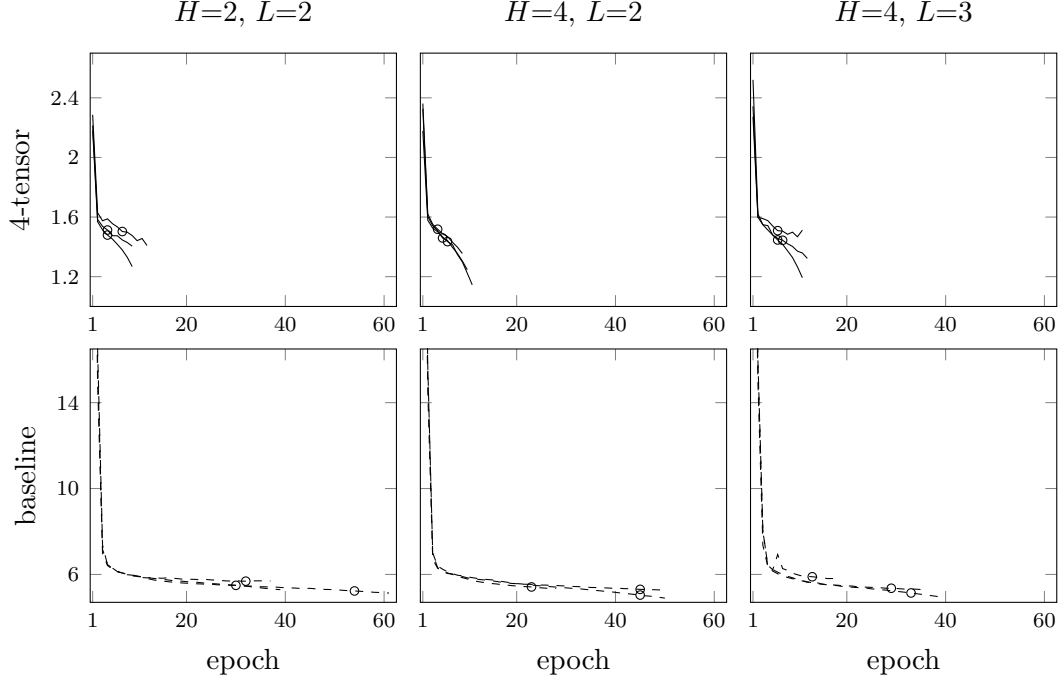

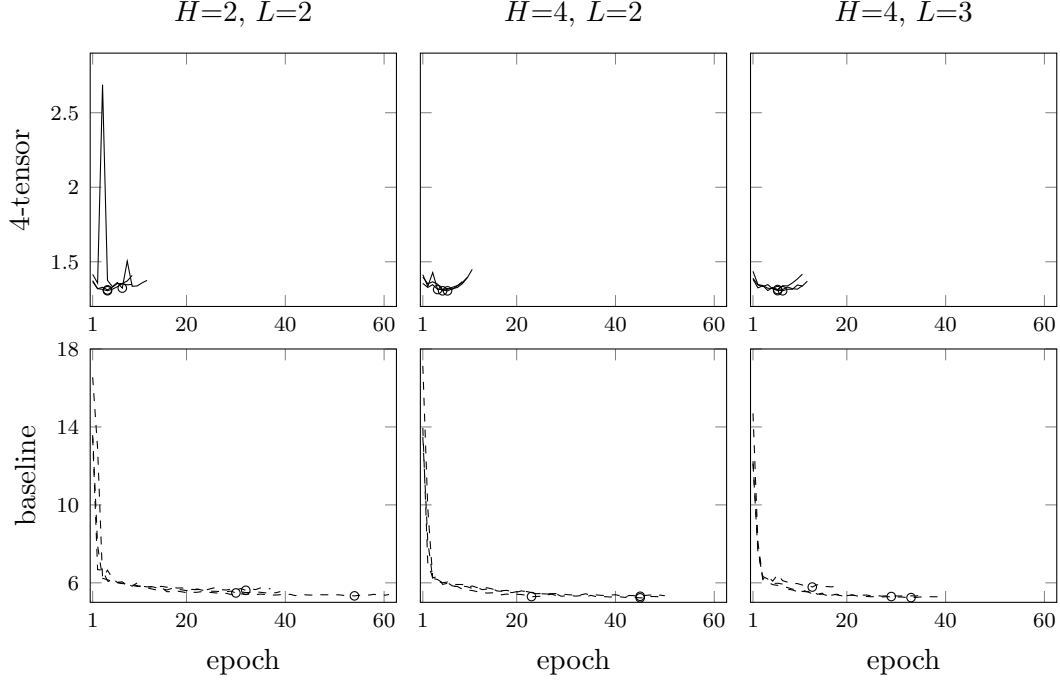
\begin{figure}[t]
\centering
\begin{tikzpicture}
\begin{groupplot}[
    group style={group size=3 by 2, horizontal sep=9pt, vertical sep=16pt},
    width=4.05cm,
    height=3.35cm,
    scale only axis,
    xmin=0.5, xmax=62.5,
    xlabel={},
    xtick={1,20,40,60},
    tick label style={font=\scriptsize},
    label style={font=\small},
    title style={font=\small},
    clip=true,
]
\nextgroupplot[ylabel={4-tensor}, title={$H{=}2$, $L{=}2$}, ymin=1.20, ymax=2.90, ytick={1.5,2.0,2.5}]
\addplot[black] table[x=epoch, y=val_ce, col sep=comma] {figures/dynamics/h2l2_s41_4t_val.csv};
\addplot[forget plot, black] table[x=epoch, y=val_ce, col sep=comma] {figures/dynamics/h2l2_s42_4t_val.csv};
\addplot[forget plot, black] table[x=epoch, y=val_ce, col sep=comma] {figures/dynamics/h2l2_s43_4t_val.csv};
\addplot[forget plot, only marks, mark=o, mark size=1.6pt, black] table[x=epoch, y=val_ce, col sep=comma] {figures/dynamics/h2l2_s41_4t_val_mark.csv};
\addplot[forget plot, only marks, mark=o, mark size=1.6pt, black] table[x=epoch, y=val_ce, col sep=comma] {figures/dynamics/h2l2_s42_4t_val_mark.csv};
\addplot[forget plot, only marks, mark=o, mark size=1.6pt, black] table[x=epoch, y=val_ce, col sep=comma] {figures/dynamics/h2l2_s43_4t_val_mark.csv};
\nextgroupplot[title={$H{=}4$, $L{=}2$}, yticklabels={}, ymin=1.20, ymax=2.90, ytick={1.5,2.0,2.5}]
\addplot[black] table[x=epoch, y=val_ce, col sep=comma] {figures/dynamics/h4l2_s41_4t_val.csv};
\addplot[forget plot, black] table[x=epoch, y=val_ce, col sep=comma] {figures/dynamics/h4l2_s42_4t_val.csv};
\addplot[forget plot, black] table[x=epoch, y=val_ce, col sep=comma] {figures/dynamics/h4l2_s43_4t_val.csv};
\addplot[forget plot, only marks, mark=o, mark size=1.6pt, black] table[x=epoch, y=val_ce, col sep=comma] {figures/dynamics/h4l2_s41_4t_val_mark.csv};
\addplot[forget plot, only marks, mark=o, mark size=1.6pt, black] table[x=epoch, y=val_ce, col sep=comma] {figures/dynamics/h4l2_s42_4t_val_mark.csv};
\addplot[forget plot, only marks, mark=o, mark size=1.6pt, black] table[x=epoch, y=val_ce, col sep=comma] {figures/dynamics/h4l2_s43_4t_val_mark.csv};
\nextgroupplot[title={$H{=}4$, $L{=}3$}, yticklabels={}, ymin=1.20, ymax=2.90, ytick={1.5,2.0,2.5}]
\addplot[black] table[x=epoch, y=val_ce, col sep=comma] {figures/dynamics/h4l3_s41_4t_val.csv};
\addplot[forget plot, black] table[x=epoch, y=val_ce, col sep=comma] {figures/dynamics/h4l3_s42_4t_val.csv};
\addplot[forget plot, black] table[x=epoch, y=val_ce, col sep=comma] {figures/dynamics/h4l3_s43_4t_val.csv};
\addplot[forget plot, only marks, mark=o, mark size=1.6pt, black] table[x=epoch, y=val_ce, col sep=comma] {figures/dynamics/h4l3_s41_4t_val_mark.csv};
\addplot[forget plot, only marks, mark=o, mark size=1.6pt, black] table[x=epoch, y=val_ce, col sep=comma] {figures/dynamics/h4l3_s42_4t_val_mark.csv};
\addplot[forget plot, only marks, mark=o, mark size=1.6pt, black] table[x=epoch, y=val_ce, col sep=comma] {figures/dynamics/h4l3_s43_4t_val_mark.csv};
\nextgroupplot[ylabel={baseline}, xlabel={epoch}, ymin=5.0, ymax=18.0, ytick={6,10,14,18}]
\addplot[black, dashed] table[x=epoch, y=val_ce, col sep=comma] {figures/dynamics/h2l2_s41_ar_val.csv};
\addplot[forget plot, black, dashed] table[x=epoch, y=val_ce, col sep=comma] {figures/dynamics/h2l2_s42_ar_val.csv};
\addplot[forget plot, black, dashed] table[x=epoch, y=val_ce, col sep=comma] {figures/dynamics/h2l2_s43_ar_val.csv};
\addplot[forget plot, only marks, mark=o, mark size=1.6pt, black] table[x=epoch, y=val_ce, col sep=comma] {figures/dynamics/h2l2_s41_ar_val_mark.csv};
\addplot[forget plot, only marks, mark=o, mark size=1.6pt, black] table[x=epoch, y=val_ce, col sep=comma] {figures/dynamics/h2l2_s42_ar_val_mark.csv};
\addplot[forget plot, only marks, mark=o, mark size=1.6pt, black] table[x=epoch, y=val_ce, col sep=comma] {figures/dynamics/h2l2_s43_ar_val_mark.csv};
\nextgroupplot[xlabel={epoch}, yticklabels={}, ymin=5.0, ymax=18.0, ytick={6,10,14,18}]
\addplot[black, dashed] table[x=epoch, y=val_ce, col sep=comma] {figures/dynamics/h4l2_s41_ar_val.csv};
\addplot[forget plot, black, dashed] table[x=epoch, y=val_ce, col sep=comma] {figures/dynamics/h4l2_s42_ar_val.csv};
\addplot[forget plot, black, dashed] table[x=epoch, y=val_ce, col sep=comma] {figures/dynamics/h4l2_s43_ar_val.csv};
\addplot[forget plot, only marks, mark=o, mark size=1.6pt, black] table[x=epoch, y=val_ce, col sep=comma] {figures/dynamics/h4l2_s41_ar_val_mark.csv};
\addplot[forget plot, only marks, mark=o, mark size=1.6pt, black] table[x=epoch, y=val_ce, col sep=comma] {figures/dynamics/h4l2_s42_ar_val_mark.csv};
\addplot[forget plot, only marks, mark=o, mark size=1.6pt, black] table[x=epoch, y=val_ce, col sep=comma] {figures/dynamics/h4l2_s43_ar_val_mark.csv};
\nextgroupplot[xlabel={epoch}, yticklabels={}, ymin=5.0, ymax=18.0, ytick={6,10,14,18}]
\addplot[black, dashed] table[x=epoch, y=val_ce, col sep=comma] {figures/dynamics/h4l3_s41_ar_val.csv};
\addplot[forget plot, black, dashed] table[x=epoch, y=val_ce, col sep=comma] {figures/dynamics/h4l3_s42_ar_val.csv};
\addplot[forget plot, black, dashed] table[x=epoch, y=val_ce, col sep=comma] {figures/dynamics/h4l3_s43_ar_val.csv};
\addplot[forget plot, only marks, mark=o, mark size=1.6pt, black] table[x=epoch, y=val_ce, col sep=comma] {figures/dynamics/h4l3_s41_ar_val_mark.csv};
\addplot[forget plot, only marks, mark=o, mark size=1.6pt, black] table[x=epoch, y=val_ce, col sep=comma] {figures/dynamics/h4l3_s42_ar_val_mark.csv};
\addplot[forget plot, only marks, mark=o, mark size=1.6pt, black] table[x=epoch, y=val_ce, col sep=comma] {figures/dynamics/h4l3_s43_ar_val_mark.csv};
\end{groupplot}
\end{tikzpicture}
\caption{Validation cross-entropy on separate axes. The upper row is the 4-tensor model on the validation split, with the same windows as \cref{fig:train}: the input is $X(3) \leftarrow \{ S(0), S(1), S(2), S(3) \}$, and the loss is the mean cross-entropy of $X(4) \leftarrow \{ S(1), S(2), S(3), S(4) \}$. The lower row is the baseline selection loss on $S(4)$, including [EOS]. Each curve is one of seeds $41$, $42$, and $43$. Open circles are the selected checkpoints in \cref{tab:runs-seeds}, the minima of these curves. Every logged epoch lies inside the plotted range. The upper-left panel includes the seed-$41$ value $2.688$ at epoch $3$. The horizontal axis is the epoch index, not wall-clock time.}
\label{fig:valid}
\end{figure}

At the selected checkpoints, the mean cross-entropy on $S(1)$, $S(2)$, and $S(3)$, recovered from the logged window loss and $\mathcal{L}_{S(4)}$ using the validation token counts, is $0.10$--$0.12$, against $5.53$--$5.60$ on $S(4)$.

\Cref{fig:dynamics} shows $\mathcal{L}_{S(4)}$ at every logged epoch of seeds $41$, $42$, and $43$. On the shared epochs inside the plotted range, the 4-tensor curve lies below the paired baseline, except at the last logged 4-tensor epoch of seed $43$ at $H{=}4$, $L{=}2$ and of seed $42$ at $H{=}4$, $L{=}3$. The open circle is the epoch of minimum selection loss. That epoch is also the minimum of $\mathcal{L}_{S(4)}$ on four of the nine 4-tensor runs. The stopping rule in \cref{sec:protocol} ends the run on the rise of the selection loss. The lowest logged baseline $\mathcal{L}_{S(4)}$ exceeds the lowest logged 4-tensor $\mathcal{L}_{S(4)}$ on every seed. The mean paired gaps of those minima are $0.303$, $0.110$, and $0.277$. \Cref{fig:train} is the loss the optimizer steps on, drawn apart from $\mathcal{L}_{S(4)}$ because the two are not on the same scale. On the 4-tensor runs the epoch-1 values are $2.227 \pm 0.054$, $2.289 \pm 0.097$, and $2.379 \pm 0.128$, and the marked checkpoints are $1.498 \pm 0.017$, $1.471 \pm 0.043$, and $1.466 \pm 0.037$, in the order of \cref{tab:runs}. $\mathcal{L}_{S(4)}$ staying near $5.5$ is the term on $S(4)$, the only sentence of $X(4)$ absent from the input $X(3)$; the training loss averages $S(4)$ with $S(1)$, $S(2)$, and $S(3)$. \Cref{fig:valid} evaluates that same pair of windows on the validation split, input $X(3)$ and target $X(4)$. Its marked points are the minima used for checkpoint selection. Across seeds those minima are $1.313 \pm 0.009$, $1.308 \pm 0.005$, and $1.308 \pm 0.005$ for the 4-tensor runs, and $5.477 \pm 0.143$, $5.278 \pm 0.043$, and $5.441 \pm 0.298$ for the baseline. On each curve the validation loss is higher after the marked epoch.

On the three matched pairs the mean last-slice cross-entropy is lower for the 4-tensor model: $5.580 \pm 0.017$ against $5.868 \pm 0.157$ at $H{=}2$, $L{=}2$, $5.545 \pm 0.010$ against $5.655 \pm 0.059$ at $H{=}4$, $L{=}2$, and $5.536 \pm 0.005$ against $5.825 \pm 0.326$ at $H{=}4$, $L{=}3$. The paired gaps are $0.288 \pm 0.165$, $0.111 \pm 0.050$, and $0.290 \pm 0.321$, and every seed's gap is positive. The $H{=}4$, $L{=}2$ pair is the closest parameter match, $172.5$M against $175.9$M, and the perplexities of the mean cross-entropies are $256$ and $286$. On the same two RTX 3080 GPUs the 4-tensor runs finished in $2.0$--$2.4$ hours and the baseline runs in $25.1$--$45.2$ hours. The $H{=}2$, $L{=}2$ pair trains a smaller 4-tensor model, $122.2$M against $175.9$M, with perplexities $265$ and $353$. The $H{=}4$, $L{=}3$ pair trains a larger 4-tensor model, $256.4$M against $226.2$M, with perplexities $254$ and $339$. The standard deviation at $H{=}4$, $L{=}3$ is carried by the baseline seed $41$, selected at epoch $13$ with cross-entropy $6.200$. The epoch budget is $50$ for seed $43$ at $H{=}2$, $L{=}2$ and at $H{=}4$, $L{=}2$, and $100$ for the other runs. Every 4-tensor checkpoint is at epoch $7$ or earlier. The baseline at $H{=}4$, $L{=}2$, seed $43$, ends at epoch $50$ because the budget is exhausted, and the selected checkpoint is epoch $45$. The baseline at $H{=}2$, $L{=}2$, seed $42$, selects epoch $54$. Each checkpoint is the minimum of that seed's selection loss on the validation split reported here.

\subsection{Rollout of $S(3)$ and $S(4)$}
\label{sec:rollout-results}

The runs in this subsection follow \cref{sec:protocol-rollout}. The window length is $T=3$, the records are the same five-sentence stories, and the reported checkpoint is the minimum of that run's selection loss. The columns are the rollout scores at that checkpoint. $S(3)$ is scored from $\{S(0), S(1), S(2)\}$, and $S(4)$ is scored from $\{S(1), S(2), \hat{S}(3)\}$. Each score is the token-weighted mean cross-entropy of the non-padding tokens of that one sentence, and the accuracy is the argmax match rate on the same tokens. The validation denominators are $37{,}582$ tokens for $S(3)$ and $32{,}173$ tokens for $S(4)$. The $H{=}2$, $L{=}2$ rows are seeds $41$, $42$, and $43$. The $H{=}4$, $L{=}2$ rows are seeds $41$ and $42$. A cell is 4-tensor / baseline. The difference on a seed is
\begin{equation}
    \Delta_{S(t)} = \mathcal{L}^{\mathrm{AR}}_{S(t)} - \mathcal{L}^{4\mathrm{T}}_{S(t)}, \qquad t \in \{3, 4\}.
\end{equation}
A positive value means the 4-tensor cross-entropy is lower. Each summary entry is the mean and the sample standard deviation of the paired differences in that row. Every paired difference is positive on both sentences.

\begin{table}[t]
\centering
\footnotesize
\setlength{\tabcolsep}{5pt}
\caption{Rollout at each seed's selected checkpoint. Each entry is 4-tensor / baseline: the mean and the sample standard deviation. The $H{=}2$, $L{=}2$ rows use seeds $41$, $42$, and $43$, with $122.2$M / $175.9$M parameters. The $H{=}4$, $L{=}2$ rows use seeds $41$ and $42$, with $172.5$M / $175.9$M parameters. $\mathcal{L}$ is the mean cross-entropy on the non-padding tokens of that sentence, $37{,}582$ for $S(3)$ and $32{,}173$ for $S(4)$, and accuracy is the argmax match rate on the same tokens. $\Delta$ is the paired difference defined in the text.}
\label{tab:rollout}
\vspace{6pt}
\begin{tabular}{@{}rrlccc@{}}
\toprule
$H$ & $L$ & sentence & $\mathcal{L}$ & acc.\ (\%) & $\Delta$ \\
\midrule
2 & 2 & $S(3)$ & $5.838 \pm 0.034$ / $6.007 \pm 0.079$ & $14.1 \pm 0.1$ / $7.0 \pm 1.8$ & $0.169 \pm 0.067$ \\
2 & 2 & $S(4)$ & $6.041 \pm 0.062$ / $6.721 \pm 0.410$ & $14.5 \pm 0.2$ / $2.0 \pm 1.5$ & $0.680 \pm 0.350$ \\
4 & 2 & $S(3)$ & $5.795 \pm 0.002$ / $5.903 \pm 0.014$ & $13.9 \pm 0.3$ / $8.6 \pm 1.2$ & $0.108 \pm 0.016$ \\
4 & 2 & $S(4)$ & $6.036 \pm 0.026$ / $7.217 \pm 0.021$ & $14.1 \pm 0.9$ / $0.6 \pm 0.1$ & $1.181 \pm 0.005$ \\
\bottomrule
\end{tabular}
\end{table}

\begin{table}[t]
\centering
\footnotesize
\setlength{\tabcolsep}{3.5pt}
\caption{Each seed of the rollout at its own selected checkpoint. The first block is $H{=}2$, $L{=}2$ and the second is $H{=}4$, $L{=}2$. Epoch is selected/last, written as 4-tensor / baseline. The other columns match \cref{tab:rollout}.}
\label{tab:rollout-seeds}
\vspace{6pt}
\begin{tabular}{@{}rrrrrrrr@{}}
\toprule
seed & epoch & $\mathcal{L}_{S(3)}$ & acc.\ $S(3)$ & $\Delta_{S(3)}$ & $\mathcal{L}_{S(4)}$ & acc.\ $S(4)$ & $\Delta_{S(4)}$ \\
\midrule
\multicolumn{8}{@{}l}{$H{=}2$, $L{=}2$} \\
41 & 4/9 / 34/40 & 5.877 / 6.048 & 14.0 / 5.1 & 0.171 & 5.983 / 6.396 & 14.4 / 3.0 & 0.413 \\
42 & 6/11 / 34/40 & 5.822 / 6.058 & 14.2 / 7.5 & 0.236 & 6.035 / 6.586 & 14.8 / 2.8 & 0.551 \\
43 & 3/8 / 36/50 & 5.815 / 5.916 & 14.0 / 8.6 & 0.101 & 6.106 / 7.182 & 14.3 / 0.3 & 1.076 \\
\midrule
\multicolumn{8}{@{}l}{$H{=}4$, $L{=}2$} \\
41 & 5/10 / 37/46 & 5.796 / 5.893 & 14.2 / 7.7 & 0.097 & 6.017 / 7.202 & 14.7 / 0.5 & 1.185 \\
42 & 5/10 / 38/50 & 5.793 / 5.913 & 13.7 / 9.5 & 0.120 & 6.054 / 7.231 & 13.4 / 0.7 & 1.177 \\
\bottomrule
\end{tabular}
\end{table}

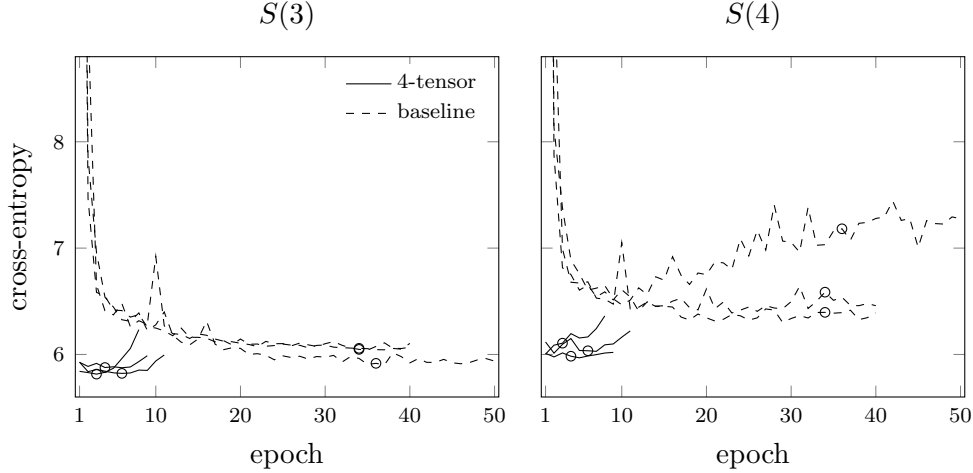
\begin{figure}[t]
\centering
\begin{tikzpicture}
\begin{groupplot}[
    group style={group size=2 by 1, horizontal sep=16pt},
    width=5.6cm,
    height=4.5cm,
    scale only axis,
    xmin=0.5, xmax=50.5,
    ymin=5.6, ymax=8.8,
    xlabel={epoch},
    xtick={1,10,20,30,40,50},
    ytick={6,7,8},
    tick label style={font=\scriptsize},
    label style={font=\small},
    title style={font=\small},
    legend style={font=\scriptsize, draw=none, at={(0.98,0.98)}, anchor=north east},
    legend cell align=left,
    clip=true,
]
\nextgroupplot[ylabel={cross-entropy}, title={$S(3)$}]
\addplot[black] table[x=epoch, y=s3_ce, col sep=comma] {figures/rollout/h2l2_s41_4t_s3.csv};
\addplot[forget plot, black] table[x=epoch, y=s3_ce, col sep=comma] {figures/rollout/h2l2_s42_4t_s3.csv};
\addplot[forget plot, black] table[x=epoch, y=s3_ce, col sep=comma] {figures/rollout/h2l2_s43_4t_s3.csv};
\addplot[black, dashed] table[x=epoch, y=s3_ce, col sep=comma] {figures/rollout/h2l2_s41_ar_s3.csv};
\addplot[forget plot, black, dashed] table[x=epoch, y=s3_ce, col sep=comma] {figures/rollout/h2l2_s42_ar_s3.csv};
\addplot[forget plot, black, dashed] table[x=epoch, y=s3_ce, col sep=comma] {figures/rollout/h2l2_s43_ar_s3.csv};
\addplot[forget plot, only marks, mark=o, mark size=1.8pt, black] table[x=epoch, y=s3_ce, col sep=comma] {figures/rollout/h2l2_s41_4t_s3_mark.csv};
\addplot[forget plot, only marks, mark=o, mark size=1.8pt, black] table[x=epoch, y=s3_ce, col sep=comma] {figures/rollout/h2l2_s42_4t_s3_mark.csv};
\addplot[forget plot, only marks, mark=o, mark size=1.8pt, black] table[x=epoch, y=s3_ce, col sep=comma] {figures/rollout/h2l2_s43_4t_s3_mark.csv};
\addplot[forget plot, only marks, mark=o, mark size=1.8pt, black] table[x=epoch, y=s3_ce, col sep=comma] {figures/rollout/h2l2_s41_ar_s3_mark.csv};
\addplot[forget plot, only marks, mark=o, mark size=1.8pt, black] table[x=epoch, y=s3_ce, col sep=comma] {figures/rollout/h2l2_s42_ar_s3_mark.csv};
\addplot[forget plot, only marks, mark=o, mark size=1.8pt, black] table[x=epoch, y=s3_ce, col sep=comma] {figures/rollout/h2l2_s43_ar_s3_mark.csv};
\legend{4-tensor, baseline}
\nextgroupplot[title={$S(4)$}, yticklabels={}]
\addplot[black] table[x=epoch, y=s4_ce, col sep=comma] {figures/rollout/h2l2_s41_4t_s4.csv};
\addplot[forget plot, black] table[x=epoch, y=s4_ce, col sep=comma] {figures/rollout/h2l2_s42_4t_s4.csv};
\addplot[forget plot, black] table[x=epoch, y=s4_ce, col sep=comma] {figures/rollout/h2l2_s43_4t_s4.csv};
\addplot[black, dashed] table[x=epoch, y=s4_ce, col sep=comma] {figures/rollout/h2l2_s41_ar_s4.csv};
\addplot[forget plot, black, dashed] table[x=epoch, y=s4_ce, col sep=comma] {figures/rollout/h2l2_s42_ar_s4.csv};
\addplot[forget plot, black, dashed] table[x=epoch, y=s4_ce, col sep=comma] {figures/rollout/h2l2_s43_ar_s4.csv};
\addplot[forget plot, only marks, mark=o, mark size=1.8pt, black] table[x=epoch, y=s4_ce, col sep=comma] {figures/rollout/h2l2_s41_4t_s4_mark.csv};
\addplot[forget plot, only marks, mark=o, mark size=1.8pt, black] table[x=epoch, y=s4_ce, col sep=comma] {figures/rollout/h2l2_s42_4t_s4_mark.csv};
\addplot[forget plot, only marks, mark=o, mark size=1.8pt, black] table[x=epoch, y=s4_ce, col sep=comma] {figures/rollout/h2l2_s43_4t_s4_mark.csv};
\addplot[forget plot, only marks, mark=o, mark size=1.8pt, black] table[x=epoch, y=s4_ce, col sep=comma] {figures/rollout/h2l2_s41_ar_s4_mark.csv};
\addplot[forget plot, only marks, mark=o, mark size=1.8pt, black] table[x=epoch, y=s4_ce, col sep=comma] {figures/rollout/h2l2_s42_ar_s4_mark.csv};
\addplot[forget plot, only marks, mark=o, mark size=1.8pt, black] table[x=epoch, y=s4_ce, col sep=comma] {figures/rollout/h2l2_s43_ar_s4_mark.csv};
\end{groupplot}
\end{tikzpicture}
\caption{Validation cross-entropy of the $T{=}3$ rollout at $H{=}2$, $L{=}2$. The left panel scores $\hat{S}(3)$ from $\{S(0), S(1), S(2)\}$. The right panel scores $\hat{S}(4)$ from $\{S(1), S(2), \hat{S}(3)\}$. Each solid curve is one 4-tensor seed and each dashed curve is the paired baseline, for seeds $41$, $42$, and $43$. Open circles are the selected checkpoints in \cref{tab:rollout-seeds}. Epoch 1 of each baseline lies above the plotted range, $15.92$--$17.07$ on $S(3)$ and $17.02$--$20.03$ on $S(4)$. Epoch 2 of seed $41$ also lies above it, at $9.15$ on $S(3)$ and $9.60$ on $S(4)$. The horizontal axis is the epoch index, not wall-clock time: the 4-tensor runs took $1.1$--$1.5$ hours and the baseline runs took $32.9$--$41.3$ hours.}
\label{fig:rollout}
\end{figure}

\begin{figure}[t]
\centering
\begin{tikzpicture}
\begin{groupplot}[
    group style={group size=2 by 1, horizontal sep=16pt},
    width=5.6cm,
    height=4.5cm,
    scale only axis,
    xmin=0.5, xmax=50.5,
    ymin=5.6, ymax=8.8,
    xlabel={epoch},
    xtick={1,10,20,30,40,50},
    ytick={6,7,8},
    tick label style={font=\scriptsize},
    label style={font=\small},
    title style={font=\small},
    legend style={font=\scriptsize, draw=none, at={(0.98,0.98)}, anchor=north east},
    legend cell align=left,
    clip=true,
]
\nextgroupplot[ylabel={cross-entropy}, title={$S(3)$}]
\addplot[black] table[x=epoch, y=s3_ce, col sep=comma] {figures/rollout/h4l2_s41_4t_s3.csv};
\addplot[forget plot, black] table[x=epoch, y=s3_ce, col sep=comma] {figures/rollout/h4l2_s42_4t_s3.csv};
\addplot[black, dashed] table[x=epoch, y=s3_ce, col sep=comma] {figures/rollout/h4l2_s41_ar_s3.csv};
\addplot[forget plot, black, dashed] table[x=epoch, y=s3_ce, col sep=comma] {figures/rollout/h4l2_s42_ar_s3.csv};
\addplot[forget plot, only marks, mark=o, mark size=1.8pt, black] table[x=epoch, y=s3_ce, col sep=comma] {figures/rollout/h4l2_s41_4t_s3_mark.csv};
\addplot[forget plot, only marks, mark=o, mark size=1.8pt, black] table[x=epoch, y=s3_ce, col sep=comma] {figures/rollout/h4l2_s42_4t_s3_mark.csv};
\addplot[forget plot, only marks, mark=o, mark size=1.8pt, black] table[x=epoch, y=s3_ce, col sep=comma] {figures/rollout/h4l2_s41_ar_s3_mark.csv};
\addplot[forget plot, only marks, mark=o, mark size=1.8pt, black] table[x=epoch, y=s3_ce, col sep=comma] {figures/rollout/h4l2_s42_ar_s3_mark.csv};
\legend{4-tensor, baseline}
\nextgroupplot[title={$S(4)$}, yticklabels={}]
\addplot[black] table[x=epoch, y=s4_ce, col sep=comma] {figures/rollout/h4l2_s41_4t_s4.csv};
\addplot[forget plot, black] table[x=epoch, y=s4_ce, col sep=comma] {figures/rollout/h4l2_s42_4t_s4.csv};
\addplot[black, dashed] table[x=epoch, y=s4_ce, col sep=comma] {figures/rollout/h4l2_s41_ar_s4.csv};
\addplot[forget plot, black, dashed] table[x=epoch, y=s4_ce, col sep=comma] {figures/rollout/h4l2_s42_ar_s4.csv};
\addplot[forget plot, only marks, mark=o, mark size=1.8pt, black] table[x=epoch, y=s4_ce, col sep=comma] {figures/rollout/h4l2_s41_4t_s4_mark.csv};
\addplot[forget plot, only marks, mark=o, mark size=1.8pt, black] table[x=epoch, y=s4_ce, col sep=comma] {figures/rollout/h4l2_s42_4t_s4_mark.csv};
\addplot[forget plot, only marks, mark=o, mark size=1.8pt, black] table[x=epoch, y=s4_ce, col sep=comma] {figures/rollout/h4l2_s41_ar_s4_mark.csv};
\addplot[forget plot, only marks, mark=o, mark size=1.8pt, black] table[x=epoch, y=s4_ce, col sep=comma] {figures/rollout/h4l2_s42_ar_s4_mark.csv};
\end{groupplot}
\end{tikzpicture}
\caption{Validation cross-entropy of the $T{=}3$ rollout at $H{=}4$, $L{=}2$, seeds $41$ and $42$. The panels and the curve styles match \cref{fig:rollout}. Epoch 1 of each baseline lies above the plotted range, $14.22$--$16.38$ on $S(3)$ and $14.57$--$17.46$ on $S(4)$. Epoch 2 of seed $41$ on $S(4)$ is $9.43$ and also lies above it. The 4-tensor runs each took $1.7$ hours and the baseline runs took $37.9$--$40.8$ hours.}
\label{fig:rollout-h4l2}
\end{figure}
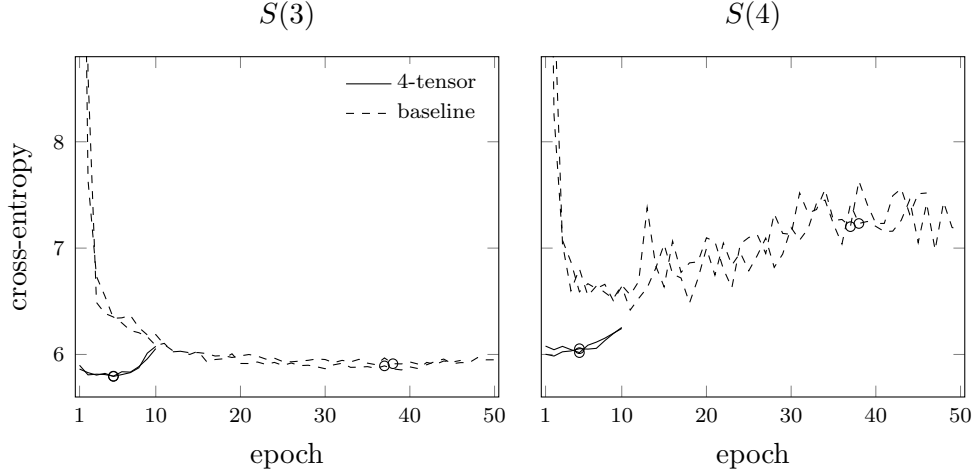

\Cref{fig:rollout,fig:rollout-h4l2} show both rollout scores at every logged epoch. On every shared epoch the 4-tensor curve lies below the paired baseline. The open circle is the epoch of minimum selection loss. For $S(3)$ that epoch is also the minimum of the plotted curve on two of the three 4-tensor runs at $H{=}2$, $L{=}2$ and on both 4-tensor runs at $H{=}4$, $L{=}2$. For $S(4)$ that epoch is the minimum of the plotted curve on the 4-tensor seed $41$ at $H{=}4$, $L{=}2$, and the selected epoch differs from that minimum on each of the other nine runs. The lowest logged baseline score exceeds the lowest logged 4-tensor score on every seed. At $H{=}2$, $L{=}2$ the mean paired gaps of those minima are $0.158$ on $S(3)$ and $0.392$ on $S(4)$, and the gaps at the selected checkpoints are $0.169 \pm 0.067$ and $0.680 \pm 0.350$. The standard deviation of that $S(4)$ gap is carried by the baseline seed $43$, selected at epoch $36$ with cross-entropy $7.182$ and accuracy $0.3\%$. That run ends at epoch $50$ because the budget is exhausted, and its lowest logged $S(4)$ is $6.467$ at epoch $10$. The perplexities of the mean cross-entropies are $343$ and $406$ on $S(3)$, and $420$ and $830$ on $S(4)$. The relative reductions are $2.8\% \pm 1.1$ and $9.9\% \pm 4.5$. Those 4-tensor runs finished in $1.1$--$1.5$ hours and the baseline runs in $32.9$--$41.3$ hours.

At $H{=}4$, $L{=}2$ the paired gaps of the trajectory minima are $0.084$ on $S(3)$ and $0.470$ on $S(4)$, and the gaps at the selected checkpoints are $0.108 \pm 0.016$ and $1.181 \pm 0.005$. Both baseline seeds are selected with $S(4)$ near $7.2$ and accuracy $0.5\%$ and $0.7\%$. Seed $42$ ends at epoch $50$ because the budget is exhausted, and the selected checkpoint is epoch $38$. Its lowest logged $S(4)$ is $6.418$ at epoch $11$. Seed $41$ stops at epoch $46$, selected at epoch $37$, and its lowest logged $S(4)$ is $6.524$ at epoch $9$. The perplexities of the mean cross-entropies are $329$ and $366$ on $S(3)$, and $418$ and $1{,}362$ on $S(4)$. The relative reductions are $1.8\% \pm 0.3$ and $16.4\% \pm 0.1$. The 4-tensor runs each finished in $1.7$ hours and the baseline runs in $37.9$--$40.8$ hours.

\section{Discussion}
\label{sec:discussion}

The rank-4 factorization forces the model to distinguish semantic content from temporal
context, keeping the $d_x$- and $d_t$-fibers as distinct tensor factors. The
separation is imposed only at the boundaries, through rank-1 input embeddings and rank-1
readout functionals, while the internal weight $W_{mn;pq}$ remains a general bilinear map
that admits arbitrary cross-fiber interaction. Coupling two representation subspaces
through a bilinear weight tensor was used in deep tensor neural networks for speech
recognition \cite{yu2013dtnn}; here the same kind of coupling parameterizes the query,
key, and value maps of an attention mechanism over a window indexed jointly by position
and time. The architecture does not, however, force
transition rules to be abstracted from the objects they apply to; it preserves a
representation in which such abstraction can emerge and be tested.

\paragraph{Future directions.}
This representation makes a further question testable: whether a learner of coupled
agent--environment dynamics can acquire, without structural annotation or
recursion-specific modules, an operation whose output fills its own argument slot. Nested
containment is a minimal setting for this question, and with identical containers the
iterated operation takes the form of Zermelo's successor construction, $0 = \varnothing$,
$n{+}1 = \{n\}$ \cite{zermelo1908untersuchungen}. We develop this program, including criteria under which it would be
refuted, in a separate document.

A multi-step rollout, with parameters held fixed, would score sentences from $S(T)$ onward under free-running inputs. For an evaluation horizon $T_{\mathrm{eval}} \ge T$,
\begin{equation}
    \mathcal{L}_{\text{eval}} = \sum_{t=T}^{T_{\text{eval}}} \sum_{x=0}^{N-1} \sum_{v \in \mathcal{V}} -y_{xv}(t) \log \hat{p}_{xv}(t),
\end{equation}
where $t$ indexes the sentence $S(t)$, $x \in \{0,\ldots,N-1\}$, and $\hat{p}_{xv}(t)$ conditions on earlier predictions rather than on the ground-truth sentence. \Cref{sec:rollout-results} reports this comparison at window length $T=3$ and horizon $T_{\mathrm{eval}}=4$. The logged scores are the token-weighted means of $S(3)$ and of $S(4)$, kept separate rather than added into the sum above. Denoising may be performed on interpolations between generated images.

\paragraph{Patent Notice}
The methods described in this paper are the subject of a pending Korean
patent application (Application No.~\mbox{10-2026-0191535}, filed
October~6, 2026) by IndigoWave Inc. Researchers interested in using
these methods are welcome to contact the authors.

\bibliographystyle{plainnat}


\end{document}